\documentclass[journal]{IEEEtran}
\usepackage{amsmath,amsfonts}
\usepackage{array}
\usepackage{textcomp}
\usepackage{stfloats}
\usepackage{float}
\usepackage{threeparttable}
\usepackage{url}
\usepackage{verbatim}
\usepackage{graphicx}
\ifCLASSOPTIONcompsoc
\usepackage[caption=false, font=normalsize, labelfont=sf, textfont=sf]{subfig}
\else
\usepackage[caption=false, font=footnotesize]{subfig}
\fi

\usepackage[linesnumbered, ruled]{algorithm2e}
\SetKwRepeat{Do}{do}{while}
\SetKwInOut{Init}{Initialization}

\usepackage{hyperref}
\usepackage{booktabs}
\usepackage{multirow}
\usepackage[table]{xcolor}
\usepackage{makecell}
\usepackage{color}
\usepackage{amssymb}
\usepackage[numbers,sort&compress]{natbib}

\newtheorem{theorem}{Theorem}[section]

\begin{document}
	%
	\title{Multi-Objective Evolutionary Optimization of Chance-Constrained Multiple-Choice Knapsack Problems with Implicit Probability Distributions}
	%
	%
	%

	\author{Xuanfeng~Li,
		Shengcai~Liu,
		Wenjie~Chen,
		Yew-Soon~Ong,
		Ke~Tang
		\thanks{Xuanfeng Li, Shengcai Liu and Ke Tang are with the Guangdong Provincial Key Laboratory of Brain-Inspired Intelligent Computation, 
			Department of Computer Science and Engineering,
			Southern University of Science and Technology, Shenzhen 518055, China (lixf2020@mail.sustech.edu.cn, liusc3@sustech.edu.cn, tangk3@sustech.edu.cn).
			
			Wenjie Chen is with the School of Information Management, Central China Normal University, Wuhan 430079, China (chenwj6@ccnu.edu.cn).
			
			Yew-Soon Ong is with the Centre for Frontier AI Research, Institute of High Performance Computing, Agency for Science Technology and Research, Singapore 138632, and also with the College of Computing and Data Science, Nanyang Technological University, Singapore 639798 (asysong@ntu.edu.sg).
			
		}
	}
	
	\maketitle
	\begin{abstract}
		The multiple-choice knapsack problem (MCKP) is a classic combinatorial optimization with wide practical applications. 
		This paper investigates a significant yet underexplored extension of MCKP: the multi-objective chance-constrained MCKP (MO-CCMCKP) under implicit probability distributions.
		The goal of the problem is to simultaneously minimize the total cost and maximize the confidence level of satisfying the capacity constraint, capturing essential trade-offs in domains like 5G network configuration. 
		To address the computational challenge of evaluating chance constraints under implicit distributions, we first propose an order-preserving efficient resource allocation Monte Carlo (OPERA-MC) method.
		This approach adaptively allocates sampling resources to preserve dominance relationships while reducing evaluation time significantly.
		Further, we develop NHILS, a hybrid evolutionary algorithm that integrates specialized initialization and local search into NSGA-II to navigate sparse feasible regions.
		Experiments on synthetic benchmarks and real-world 5G network configuration benchmarks demonstrate that NHILS consistently outperforms several state-of-the-art multi-objective optimizers in convergence, diversity, and feasibility.
		The benchmark instances and source code will be made publicly available to facilitate research in this area.
	\end{abstract}
	
	\begin{IEEEkeywords}
		Multi-objective knapsack problems, chance constrains, evolutionary algorithm, real-world applications
	\end{IEEEkeywords}
	
	\IEEEpeerreviewmaketitle
	
	\section{Introduction}
	\label{chap1:introduction}
	
	\IEEEPARstart{T}{he} multiple-choice knapsack problem (MCKP) is a fundamental combinatorial optimization problem that has attracted significant attention within the computational intelligence community due to its diverse applications, including supply chain design~\cite{sharkey2011class}, 5G network configuration~\cite{nasrallah2018ultra}, target market selection \cite{taaffe2008target}, and financial portfolio management~\cite{nauss19780}.
	While exact and heuristic algorithms for the deterministic MCKP are well-established~\cite{dyer1984branch, dudzinski1987exact, hifi2006reactive}, they typically assume that item weights are fixed and known.
	In many real-world scenarios, however, item weights are uncertain due to variable resource consumption or fluctuating delays.
	Relying on deterministic assumptions in these cases often leads to solutions that are either suboptimal or infeasible in practice.
	
\begin{figure*}[!t]
		\centering
		\includegraphics[width=0.95\textwidth]{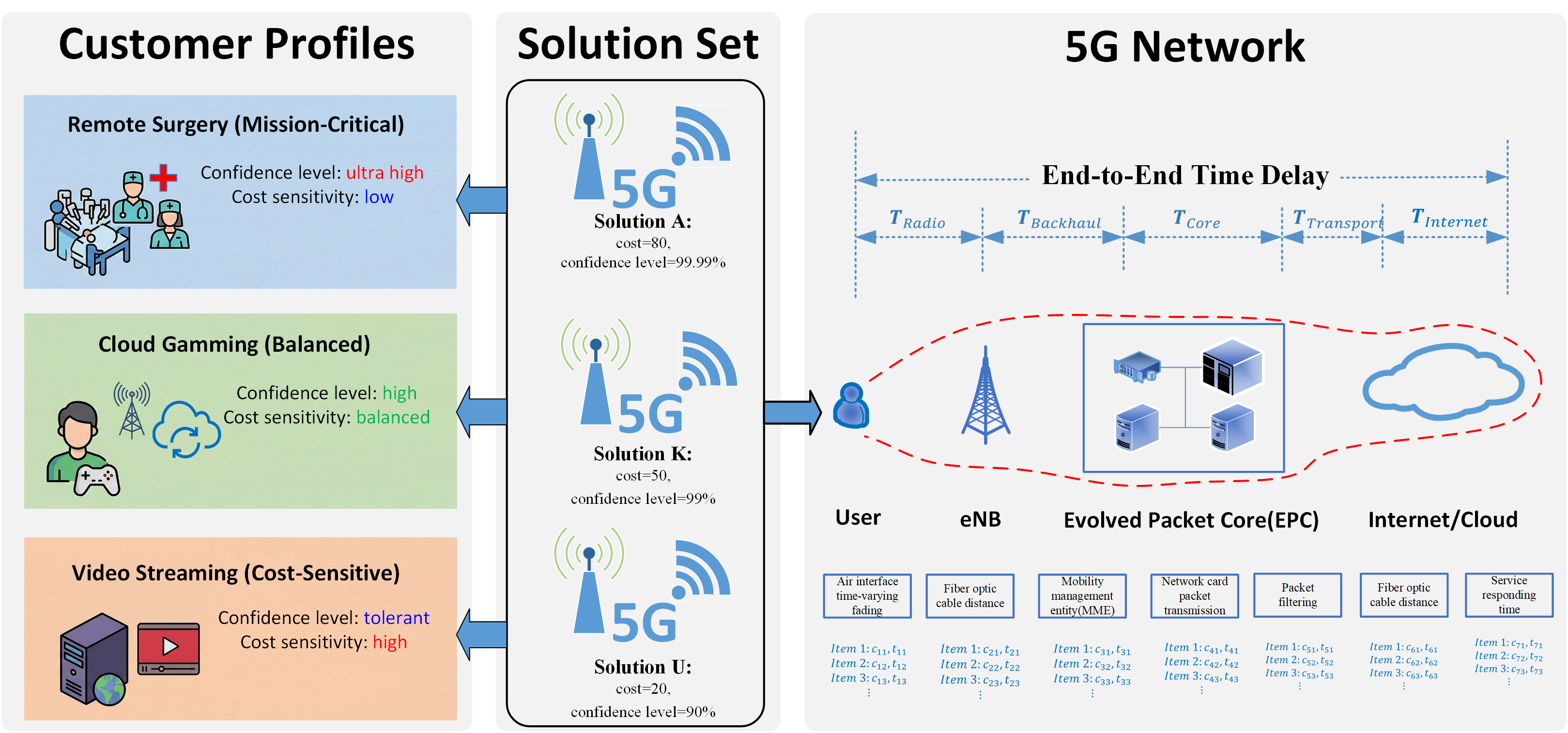}
		\caption{Technical architectures and requirement spectrum of heterogeneous 5G network service scenarios: Remote Surgery (mission-critical); Cloud Gaming (balanced); and Video Streaming (cost-sensitive).}
		\label{fig:5G-multiobjective}
	\end{figure*}

	Specifically, fluctuating item weights can easily cause a chosen solution to exceed the knapsack's capacity.
	To manage this risk, chance-constrained (CC) programming offers a probabilistic framework.
	Instead of requiring strict constraint satisfaction, CC ensures that capacity limits are met with at least a user-defined probability.
	In the resulting chance-constrained MCKP (CCMCKP) model, item weights are treated as random variables.
	Yet, most existing studies on stochastic knapsack problems assume these variables follow known probability distributions, such as Gaussian or uniform distributions~\cite{assimi2020evolutionary, hewa2024using, perera2024multi}. 
	This assumption allows for analytical reformulations or simplified sampling.
	However, it fails to model complex systems where distributions are implicit and cannot be expressed in closed form. 
	A notable example is the 5G network configuration~\cite{nasrallah2018ultra} (see Figure~\ref{fig:5G-multiobjective}), where transmission delays result from complex interactions and must be estimated via measurement or simulation rather than derived analytically.
	
\begin{figure}[htbp]
		\centering
		\includegraphics[width=0.48\textwidth]{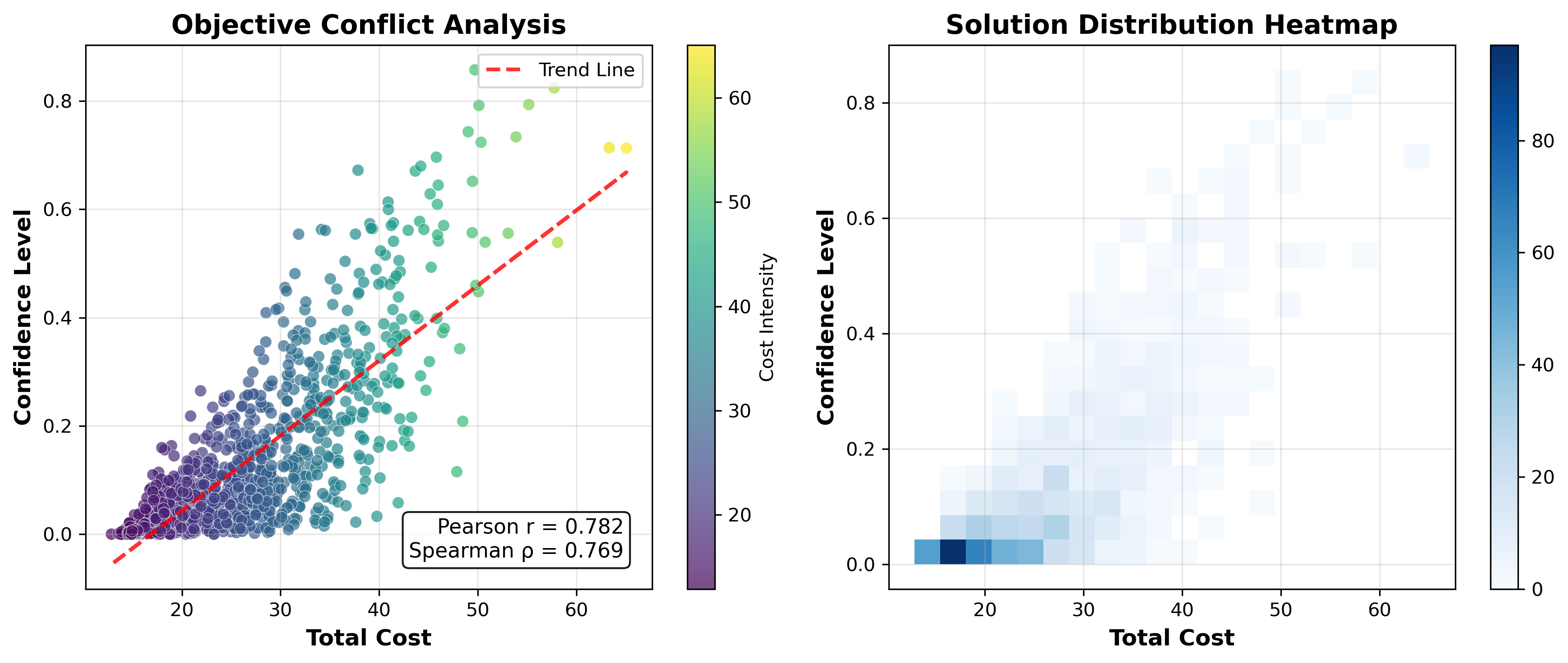}
		\caption{Trade-off and solution distribution heatmap between cost and CL.}
		\label{fig:objective_conflict}
	\end{figure}

	Beyond implicit uncertainty, practical decision-making often involves balancing conflicting objectives. 
	This necessitates a transition from single-objective optimization to a multi-objective framework. 
	In the 5G network configuration scenario shown in Figure~\ref{fig:5G-multiobjective}, user requirements span a ``performance--cost'' spectrum. 
	Mission-critical applications, such as remote surgery, require high reliability regardless of cost, whereas services like video streaming prioritize cost reduction. 
	Intermediate services, such as cloud gaming, require a balance between these two. 
	To confirm this conflict, we analyzed the correlation between total cost and confidence level (reliability) using the APP-ls1 instance proposed by~\cite{Li2024chance}, which is a real-world instance from the 5G network configuration application of a Chinese telecommunications company.
	As shown in Figure~\ref{fig:objective_conflict}, our analysis reveals a strong negative correlation (Pearson’s $r = -0.77$, $p < 0.001$), confirming that cost minimization and reliability maximization are competing objectives. 
	This conflict motivates the formulation of the multi-objective CCMCKP (MO-CCMCKP) in this work.
	
	In the literature, the MO-CCMCKP remains underexplored, as it resides at the challenging intersection of multi-objective optimization, implicit chance constraints, and combinatorial structures.
	While multi-objective Evolutionary Algorithms (MOEAs) have progressed a lot recently~\cite{lin2019clustering,tian2022evolutionary}, standard MOEAs struggle with MO-CCMCKP for three reasons: 
	(i) evaluating chance constraints via sampling is time-consuming;
	(ii) implicit distributions make it difficult to verify Pareto dominance; 
	and (iii) the multiple-choice structure requires specialized operators to maintain feasibility.
	
	To address these challenges, this paper proposes a novel MOEA for solving the MO-CCMCKP with implicit probability distributions.
	Our main contributions are summarized below.
	\begin{enumerate}
		\item We formally define the MO-CCMCKP under implicit probability distributions. 
		Unlike traditional models using closed-form distributions, our formulation uses sampling to capture real-world uncertainties, making it applicable to domains like 5G network configuration.
		
		\item We propose the order-preserving efficient resource allocation Monte Carlo (OPERA-MC) method for evaluating solutions.
		To resolve the computational bottleneck of sampling-based evaluation, OPERA-MC adaptively allocates sampling resources. 
		It drastically reduces evaluation time while preserving dominance relationships, providing a scalable solution approach for implicit chance-constrained problems.
		
		\item We develop the NSGA-II with hybrid initialization and local search (NHILS) for solving MO-CCMCKP.
		This algorithm integrates a hybrid initialization strategy (blending greedy and random approaches) with probabilistically triggered local search operators. 
		This design effectively balances exploration with the exploitation of high-quality feasible regions within the combinatorial search space. 
		
		\item We conduct extensive experiments on synthetic benchmarks and real-world application benchmark instances. 
		The results show that NHILS outperforms several state-of-the-art multi-objective optimizers in convergence, diversity, and feasibility. 
		Ablation studies further validate the efficacy of the proposed algorithmic components.
	\end{enumerate}
	
	The rest of this paper is organized as follows.
	Section~\ref{sec:related_work} reviews related work. 
	Section~\ref{sec:problem_formulation} defines the problem. 
	Section~\ref{sec:evaluation_method} and Section~\ref{sec:algorithm} detail the proposed algorithm. 
	Section~\ref{sec:experiments} presents the experiments. 
	Finally, Section~\ref{sec:conclusion} concludes the paper and outlines future directions.
	
	\section{Related Work}
	\label{sec:related_work}
	This section reviews the methodological foundations and recent advances relevant to the MO-CCMCKP.
	We organize the review into four parts: (i) algorithmic foundations for the deterministic MCKP, (ii) the evolution of chance-constrained (CC) optimization, (iii) multi-objective extensions under uncertainty, and (iv) the capabilities and limitations of existing MOEAs.
	Finally, we identify the research gap that motivates our study.
	
	\subsection{The Deterministic MCKP}
	The deterministic MCKP is a classic combinatorial optimization problem.
	Exact methods, such as branch-and-bound~\cite{dyer1984branch} and dynamic programming~\cite{dudzinski1987exact}, are effective for small to medium instances.
	For larger problems, Pisinger~\cite{pisinger1995minimal} developed an algorithm integrating linear relaxation with greedy selection, while others proposed approximation schemes~\cite{gens1998approximate, he2016improved}.
	Metaheuristics have also been widely applied: Hifi et al.~\cite{hifi2006reactive} introduced a reactive local search, Mkaouar et al.~\cite{mkaouar2020solving} used artificial bee colony optimization, and Lamanna et al.~\cite{lamanna2022two} employed kernel search for multidimensional variants.
	While these methods are efficient in deterministic settings, they cannot accommodate stochastic variables or probabilistic constraints.
	
	\subsection{Chance-Constrained Optimization Paradigms}
	CC programming provides a formal framework for decision-making under uncertainty~\cite{shapiro2021lectures}.
	Early work on stochastic knapsack problems (KPs) focused on parametric assumptions.
	For instance, PTAS were developed for Poisson~\cite{goel1999stochastic} and Gaussian~\cite{goyal2010ptas, shabtai2018relaxed} weight distributions.
	
	When distributions are unknown, researchers typically use sample average approximation (SAA) or distributionally robust optimization (DRO).
	Mohajerin Esfahani and Kuhn~\cite{mohajerin2018data} showed that DRO over Wasserstein ambiguity sets can be reformulated into tractable convex problems.
	This has been extended to multidimensional KPs~\cite{ji2021data} and linked to conditional value-at-risk constraints~\cite{xie2021distributionally, gao2022distributionally}.
	However, SAA requires very large sample sizes to ensure feasibility under tight constraints, while DRO often produces overly conservative solutions or becomes computationally intractable for combinatorial problems.
	To overcome these limitations, Li et al.~\cite{Li2024chance} introduced the single-objective CCMCKP with implicit distributions and proposed the DDALS algorithm, which evaluates constraint satisfaction directly via sampling without parametric assumptions.
	However, that work optimizes solely for cost minimization and does not consider the multi-objective trade-offs common in practical applications.
	
	\subsection{Multi-Objective Chance-Constrained Knapsack Problems}
	Although the MO-CCMCKP remains unexplored, related stochastic knapsack variants exist.
	Perera et al.~\cite{perera2024multi} studied knapsack problems with stochastic profits, while others investigated chance-constrained submodular optimization~\cite{yan2024sliding} and sliding-window selection~\cite{kokila2024multi}.
	Assimi et al.~\cite{assimi2020evolutionary} formulated a bi-objective dynamic CCKP, maximizing constraint satisfaction probability while minimizing required capacity. 
	Hewa et al.~\cite{hewa2024using} extended this to three objectives (i.e., profit, weight, and risk) under uniform distributions, using analytical violation probabilities to approximate the Pareto front.
	However, all of these studies assume known parametric distributions, which are often unavailable in real-world systems with implicit uncertainty.
	
	\subsection{MOEAs and the Sparsity-of-Feasibility Challenge}
	
	The broader field of MOEAs has matured significantly, offering sophisticated tools for complex landscapes~\cite{tian2022evolutionary,feng2022solving,hong2024multi,song2024balancing,liu2024effective}. 
	Among these, NSGA-II remains a typical cornerstone, inspiring numerous specialized enhancements. 
	To handle dynamic environments, the D-NSGA-II variant incorporating a gradient-based local search strategy has been developed for multi-objective optimization problems~\cite{ding2019dynamic}.
	For combinatorial optimization domains, learning-based improvement methods have been integrated into NSGA-II frameworks to enhance search direction guidance and solution quality~\cite{ye2025solving}. 
	Additionally, a many-objective variant, MaNSGA-II, modifies the dominance relation and diversity maintenance mechanism to effectively handle problems with more than three objectives~\cite{mansgaii2025}.
	More recently, altruistic resource allocation has been proposed to better balance convergence and diversity~\cite{chen2025altruistic}. 
	
	Despite these developments, standard MOEAs struggle with the MO-CCMCKP due to the ``sparsity of feasibility'' phenomenon.
	In high-dimensional 5G configurations under tight chance constraints, the feasible region often constitutes a negligible fraction of the combinatorial search space.
	Consequently, random initialization and generic genetic operators often fail to locate feasible solutions, leading to stagnated search in infeasible regions.
	This necessitates a hybrid framework that couples problem-specific initialization to ``anchor'' the population in feasible regions with local search to navigate the narrow ridges of the Pareto front.
	
	In summary, the literature review reveals a gap at the intersection of (i) multiple-choice combinatorial structures, (ii) multi-objective optimization, and (iii) implicit uncertainty.
	Our work addresses this by combining sample-based evaluation with an enhanced NSGA-II backbone tailored for the MO-CCMCKP.

	\section{Problem Formulation}
	\label{sec:problem_formulation}
	This section formally defines the multi-objective chance-constrained multiple-choice knapsack problem (MO-CCMCKP).
	The MO-CCMCKP extends the classical MCKP by incorporating two conflicting objectives and chance constraints.
	A key characteristic of this formulation is that the probability distributions of item weights are implicit, i.e., they lack a closed-form expression and can only be sampled.
	
	\subsection{Mathematical Model}
	Consider a system with $m$ mutually disjoint classes $N_1, \ldots, N_m$.
	Each class $N_i$ contains a finite set of items.
	Each item $j \in N_i$ is associated with a fixed cost $c_{ij}$ and a weight $w_{ij}$, where $w_{ij}$ is a random variable with an implicit distribution.
	We assume that item weights are mutually independent across all classes and items.
	Although the probability density or mass functions of the weights are unknown, we can obtain independent and identically distributed (i.i.d.) samples from them via a simulator or empirical measurements.
	Let $W$ denote the knapsack capacity and $P_0$ represent the required confidence level (CL).
	The total weight of a solution is a random variable defined as:
	\begin{equation}
		g(\mathbf{x}) = \sum_{i=1}^{m} \sum_{j \in N_i} w_{ij} x_{ij}.
	\end{equation}
	
	The MO-CCMCKP seeks to minimize the total cost while maximizing the probability that the total weight remains within capacity (the CL).
	Mathematically, the problem is formulated as:
	\begin{subequations}
		\begin{align}
			& f_1(\mathbf{x}) = \min \sum_{i=1}^{m} \sum_{j \in N_i} c_{ij} x_{ij}, \label{eq:objective1} \\
			& f_2(\mathbf{x}) = \min -\mathcal{P}\left(g(\mathbf{x}) \leq W\right), \label{eq:objective2} \\\\
			\text{s.t.}\quad &\mathcal{P}(g(\mathbf{x}) \leq W) \geq P_0, \label{eq:chance_constraint} \\
			& \sum_{j \in N_i} x_{ij} = 1, \quad \forall i \in \{1,\ldots,m\}, \label{eq:multiple_choice} \\
			& x_{ij} \in \{0, 1\}, \quad \forall i \in \{1,\ldots,m\}, \forall j \in N_i. \label{eq:binary}
		\end{align}
	\end{subequations}
	
	Equation~\eqref{eq:objective1} represents the total cost to be minimized, while Equation~\eqref{eq:objective2} transforms the maximization of CL into a minimization problem by taking the negative value. 
	The chance constraint~\eqref{eq:chance_constraint} ensures that the probability of the total weight not exceeding the capacity $W$ is at least $P_0$. 
	Constraints~\eqref{eq:multiple_choice} and~\eqref{eq:binary} enforce the multiple‑choice structure.
	
	A primary challenge in this formulation is that the probability $\mathcal{P}\left(g(\mathbf{x}) \leq W\right)$ cannot be computed analytically.
	Instead, we have access to a sampling oracle.
	Consequently, the feasibility of a solution and the value of $f_2$ must be estimated by sampling.
	
	\subsection{Solution Representation and Fitness Evaluation}
	To solve MO‑CCMCKP with EAs, a suitable solution encoding and an efficient evaluation procedure are essential.
	
	\subsubsection{Solution Encoding}
	A candidate solution is encoded as an integer vector $\mathbf{x} = (x_1, x_2, \ldots, x_m)$, where each element $x_i$ represents the index of the selected item from class $N_i$.
	This representation inherently satisfies the multiple-choice constraints~\eqref{eq:multiple_choice} and~\eqref{eq:binary}.
	
	\subsubsection{Fitness Evaluation}
	Evaluating a solution requires computing two objective values:
	
	\begin{itemize}
		\item \textbf{Total Cost}: The first objective $f_1(\mathbf{x})$ is deterministic and calculated directly using Equation~\eqref{eq:objective1}.
		
		\item \textbf{Confidence Level (CL)}: Since weight distributions are implicit, $\mathcal{P}\left(g(\mathbf{x}) \leq W\right)$ is estimated empirically. Given $L$ i.i.d. samples of the weight vector, let $w_{i,x_i}^{(l)}$ be the $l$-th sample weight of the item selected in class $i$. The empirical confidence is:
		\begin{equation}
			\hat{P}(\mathbf{x}) = \frac{1}{L} \sum_{l=1}^{L} \mathbf{1}\left\{\sum_{i=1}^{m} w_{i,x_i}^{(l)} \leq W\right\},
		\end{equation}
		where $\mathbf{1}\{\cdot\}$ is the indicator function. The second objective is then $f_2(\mathbf{x}) = -\hat{P}(\mathbf{x})$.
	\end{itemize}
	
		

	The estimator $\hat{P}(\mathbf{x})$ is unbiased and converges to the true confidence level as $L \to \infty$.
	However, large values of $L$ significantly increase the computational burden.
	The MO-CCMCKP is NP-hard, as it generalizes the classical MCKP~\cite{sinha1979multiple}.
	The combinatorial search space of size $\prod_{i=1}^m |N_i|$, combined with the requirements of sampling-based evaluation, necessitates the specialized hybrid algorithm and efficient evaluation scheme described in the following sections. 
	
	
	
	
	\section{OPERA-MC: Efficient Evaluation of Implicit Chance Constraints}
	\label{sec:evaluation_method}
	
	Estimating solution feasibility under chance constraints is a significant computational challenge.
	Since implicit probability distributions lack closed-form expressions, the probability $\mathcal{P}\left(g(\mathbf{x}) \leq W\right)$ must be estimated via Monte Carlo (MC) simulation using samples from the underlying stochastic process.
	However, standard MC evaluation is computationally expensive when integrated into an EA that evaluates thousands of candidates. 
	Furthermore, Hoeffding's inequality indicates that the estimation error decreases at a rate of $O(1/\sqrt{N})$, meaning a ten-fold improvement in precision requires a hundred-fold increase in the number of samples $N$.
	
	To address this, we propose the order-preserving efficient resource allocation Monte Carlo (OPERA-MC) method.
	OPERA-MC adaptively allocates simulation samples based on the estimated quality of each solution.
	This preserves correct dominance relationships while reducing the total number of simulations.
	This approach is particularly suited for optimization problems where the probability distributions are implicit and exact evaluation is time-consuming.
	
	\subsection{Motivation and Problem Setting}
	Consider a general chance-constrained optimization problem with an implicit probability distribution. 
	We assume access to a \textit{sampling oracle} that generates i.i.d. samples of the random variables, but we have no analytical expression for the distribution. 
	For any candidate solution $\mathbf{x}$, we estimate the probability $p(\mathbf{x}) = \mathcal{P}\left(g(\mathbf{x}) \leq W\right)$ using the empirical mean:
	\begin{equation}
		\hat{p}_N(\mathbf{x}) = \frac{1}{N} \sum_{k=1}^{N} \mathbb{I}\left(g^{(k)}(\mathbf{x}) \leq W\right),
	\end{equation}
	where $g^{(k)}(\mathbf{x})$ is the $k$-th sample of the total weight.
	
	Multi-objective optimization requires comparing solutions based on both cost $f_1(\mathbf{x})$ and confidence $p(\mathbf{x})$.
	A solution $\mathbf{x}_A$ dominates $\mathbf{x}_B$ if $f_1(\mathbf{x}_A) \leq f_1(\mathbf{x}_B)$ and $p(\mathbf{x}_A) \geq p(\mathbf{x}_B)$, with at least one strict inequality.
	Therefore, accurately estimating $p(\mathbf{x})$ is essential for maintaining correct dominance relationships during the search process.
	
	\subsection{The OPERA-MC Algorithm}
	The core principle of OPERA-MC is that solutions with low true confidence can be identified as inferior using few samples, while high-confidence solutions require more precise evaluation to distinguish subtle differences.
	Let $T_k$ and $P_k$ denote the cumulative sample size and the confidence threshold at stage $k$, respectively, where $T_0 = 0 < T_1 < T_2 < \cdots < T_K$ and $P_0 \leq P_1 < P_2 < \cdots < P_{K-1} < P_K = \infty$.
	A solution that satisfies the threshold at stage $k$ (i.e., $\hat{p}_{T_k}(\mathbf{x}) \geq P_k$) proceeds to the next stage for more granular evaluation.
	Solutions that fail to meet the threshold are rejected early, saving computational resources.
	

	\begin{algorithm}[t]
		\caption{OPERA-MC}
		\label{alg:adaptive_mc}
		\KwIn{Solution $\mathbf{x}$, stage parameters $\{(T_k, P_k)\}_{k=1}^{K}$}
		\KwOut{Estimated CL $\hat{p}(\mathbf{x})$}
		$N_{\text{total}} \gets 0$ \\
		$\hat{p} \gets 0$ \\
		\For{$k = 1$ \KwTo $K$}{
			Perform $\Delta T_k = T_k - T_{k-1}$ additional MC samples; \\
			$N_{\text{total}} \gets N_{\text{total}} + \Delta T_k$; \\
			Update $\hat{p}$ using all $N_{\text{total}}$ samples; \\
			\If{$\hat{p} < P_k$}{
				\Return{$\hat{p}$} \tcp*{Early stop}
			}
		}
		\Return{$\hat{p}$} \tcp*{Full evaluation completed}
	\end{algorithm}
	
	\subsection{Theoretical Analysis}
	The design of OPERA-MC is supported by concentration inequalities \cite{boucheron2013concentration} that guarantee order-preservation with high probability.
	Let $X \in \{0,1\}$ be a Bernoulli random variable indicating whether a solution satisfies the weight constraint, where $\mathbb{E}[X] = p$.
	For $N$ i.i.d. samples, the empirical mean $\hat{p}_N$ follows Hoeffding's inequality:
	\begin{equation}
		\mathbb{P}\!\left(|\hat{p}_N - p| \geq \epsilon\right) \leq 2\exp(-2N\epsilon^2).
	\end{equation}
	
	Based on this, we derive the following theorem to bound the probability of erroneous dominance comparisons:
	\begin{theorem}[Pair-wise Order-Preservation Error Bound]
		\label{thm:order_preserve}
		Let $\mathbf{x}_A$ and $\mathbf{x}_B$ be two solutions with true confidence levels $p_A > p_B$. If $\mathbf{x}_A$ receives $N_A$ samples and $\mathbf{x}_B$ receives $N_B \leq N_A$ samples under OPERA-MC, the probability of erroneously concluding that $\hat{p}_B \geq \hat{p}_A$ satisfies:
		\begin{equation}
			\begin{aligned}
				\mathbb{P}\bigl(\hat{p}_B \geq \hat{p}_A\bigr) &\leq \exp\left[ - \frac{2(p_A-p_B)^2}{\frac{1}{N_B}+\frac{1}{N_A}} \right] \\
				&\leq \exp\left[ - N_B (p_A-p_B)^2 \right].
			\end{aligned}
		\end{equation}
	\end{theorem}
	
	
	The proof (provided in the Supplementary Material) applies Hoeffding's inequality to both estimators to bound the probability of the joint event $\hat{p}_B \geq \hat{p}_A$.
	This theorem confirms that large confidence gaps ($p_A - p_B$) allow for reliable early stopping with few samples, while smaller gaps require more samples.
	Consequently, OPERA-MC automatically focuses computational effort on solutions that are difficult to distinguish, ensuring both efficiency and accuracy.
	
	\subsection{Methodological Generality}
	OPERA-MC is not limited to MO-CCMCKP. 
	This approach applies to any chance-constrained optimization problem with implicit probability distributions where:
	\begin{itemize}
		\item The constraint function can be evaluated via sampling the random variables.
		\item The decision-maker compares solutions based on both deterministic objectives and constraint satisfaction probabilities.
	\end{itemize}
	This includes a wide range of stochastic problems in finance, logistics, and engineering~\cite{zhang2025liner, gassmann2012stochastic}.
	
	The order-preserving property of OPERA-MC establishes a theoretically sound foundation for integrating adaptive solution evaluation into evolutionary search.
	This efficiency gain enables the NHILS algorithm presented in the next section, where reduced evaluation overhead allows for hybrid initialization and local search within a practical computational budget.
	

	\section{The NHILS Algorithm}
	\label{sec:algorithm}
	This section presens NHILS, a MOEA tailored for the MO-CCMCKP.
	NHILS addresses the challenge of navigating a combinatorial search space where the feasible region is extremely sparse.
	Built upon the NSGA-II framework \cite{Deb2002fast}, the algorithm introduces two primary components:
	(i) a hybrid initialization strategy that anchors the population in feasible regions, and (ii) probabilistically triggered local search operators for neighborhood exploration.
	These enhancements balance global exploration with local exploitation of high-quality solutions.
	
	\subsection{Algorithmic Framework}
\begin{figure}[t]
		\centering
		\includegraphics[width=0.45\textwidth]{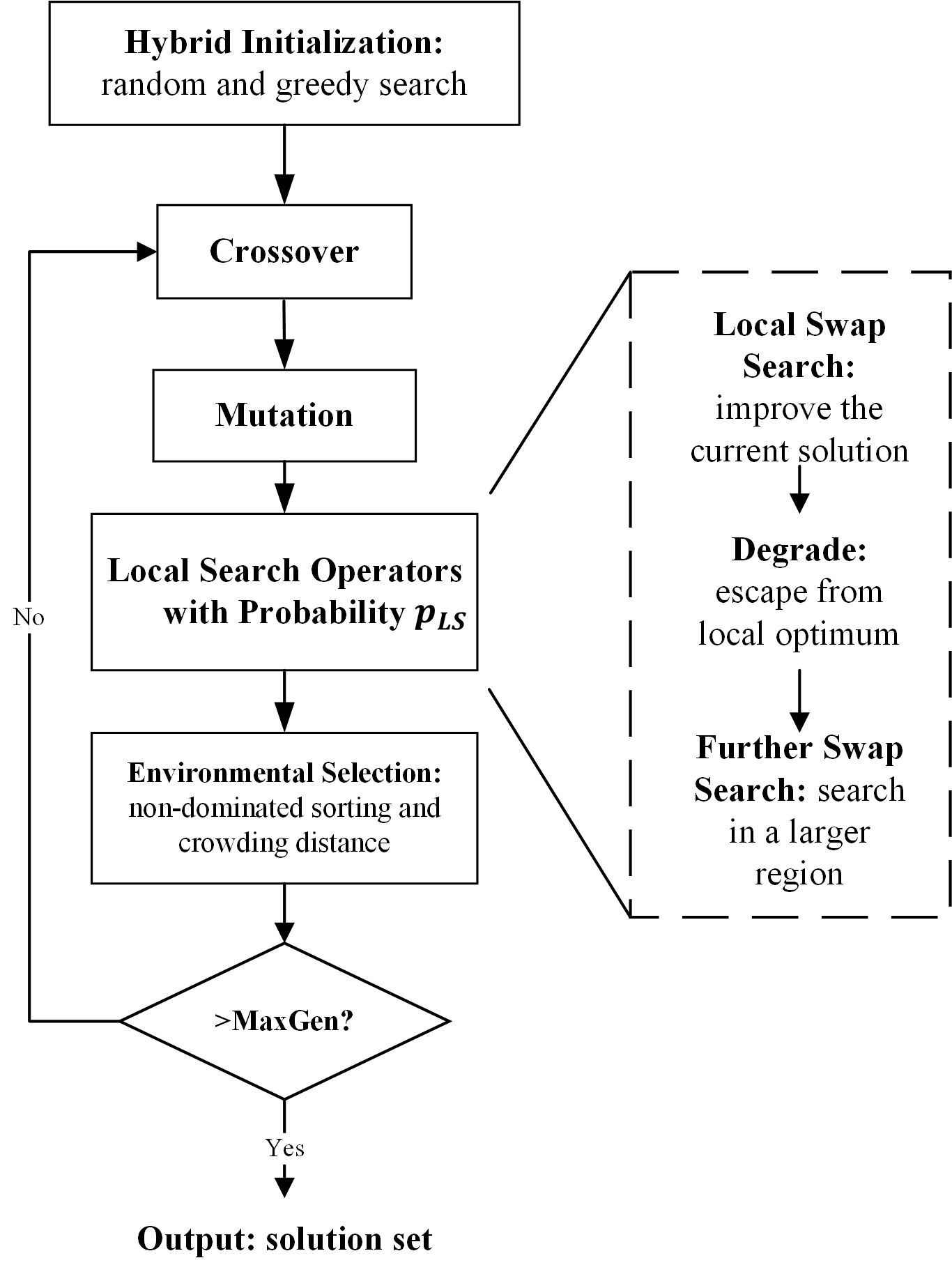}
		\caption{The block diagram of NHILS} 
		\label{fig:NHILS}
	\end{figure}

	As shown in Figure~\ref{fig:NHILS}, NHILS follows the standard NSGA-II structure while integrates problem-specific enhancements at critical stages.
	The algorithm begins with a hybrid initialization procedure to generate a diverse, high-quality initial population.
	In each generation, offspring are produced through crossover and mutation.
	Local search operators are then applied probabilistically to selected individuals.
	The merged population is evaluated using the OPERA-MC method (Section~\ref{sec:evaluation_method}) and undergoes environmental selection based on non-dominated sorting and crowding distance to form the next generation.
	
	\begin{algorithm}[tbp]
		\caption{The NHILS Algorithm}
		\label{alg:nsga2_hils}
		\SetAlgoLined
		\KwIn{Population size $S$, maximum generations $MaxGen$, local search probability $p_{LS}$}
		\KwOut{Approximated Pareto set}
		$P_0 \gets \text{HybridInitialization}(S)$\;
		$t \gets 0$\;
		\While{$t \leq MaxGen$}{
			$Q_t \gets \text{CrossoverAndMutation}(P_t)$\;
			$R_t \gets P_t \cup Q_t$\;
			\ForEach{$\mathbf{x} \in R_t$}{
				\If{rand() $< p_{LS}$}{
					$\mathbf{x}' \gets \text{LocalSearch}(\mathbf{x})$\;
					$R_t \gets R_t \cup \{\mathbf{x}'\}$\;
				}
			}
			Evaluate $R_t$ using OPERA-MC\;
			$P_{t+1} \gets \text{EnvironmentalSelection}(R_t, S)$\;
			$t \gets t+1$\;
		}
		\Return{$P_{MaxGen}$}
	\end{algorithm}
	
	
	\subsection{Surrogate Weight Metric}
	\label{subsec:surrogate_weight}
	To guide the greedy initialization and local search, we define a surrogate weight for each item that combines its mean and standard deviation:
	\begin{equation}
		\tilde{w}_{ij} = \mu_{ij} + \lambda \sigma_{ij},
	\end{equation}
	where $\mu_{ij}$ and $\sigma_{ij}$ are the sample mean and standard deviation of the weight $w_{ij}$, and $\lambda$ is the weighting parameter. 
	The parameter $\lambda$ controls the risk-aversion level; following~\cite{Li2024chance}, we set $\lambda = 3$ to emphasize reliability under uncertainty.
	
	\subsection{Hybrid Initialization}
	We employ a two-stage hybrid initialization to ensure the population starts near feasible, high-quality regions of the search space.
	In the first stage, a \textit{seed solution} is constructed.
	For each class $i$, we greedily select the item with the highest utility-to-surrogate-weight ratio: $(\max_{k \in N_i} \{c_{ik}\} - c_{ij}) / \tilde{w}_{ij}$.
	If the total surrogate weight exceeds capacity $W$, we iteratively swap items for lower-weight alternatives within their respective classes until feasibility is achieved.
	This yields a baseline feasible solution with low cost.
	
	In the second stage, we populate the remaining $S-1$ individuals through controlled perturbation of the seed individual.
	For each individual, a random subset of classes is selected, and their current items are replaced with random alternatives.
	A perturbed solution is accepted only if its total surrogate weight satisfies the capacity constraint.
	This process repeats until the population is full or the maximum perturbation attempts are reached, ensuring both feasibility and genetic diversity.
	
	
	\subsection{Local Search Operators}
	Local search is applied to individuals with probability $p_{LS}$ to explore three distinct neighborhood structures:
	\begin{itemize}
		\item \textbf{Single-Item Swap:} For each class $i$, the operator tests replacing the current item with every other item in the same class.
		If a swap maintains feasibility (based on surrogate weights) and improves at least one objective without degrading the other, it is accepted.
		
		\item \textbf{Double-Item Swap:} To escape local optima that single-item swaps cannot resolve, this operator examines simultaneous changes in two distinct classes $(i_1, i_2)$.
		It tests all item combinations from these classes and accepts improving, feasible swaps.
		
		\item \textbf{Degradation:} As a diversification mechanism, this operator randomly replaces an item in a randomly selected class. This move is accepted if the resulting solution remains feasible, even if the cost increases, to help the search move across narrow feasible ridges.
	\end{itemize}
	
	
	
	
	\subsection{Complexity Analysis}
	The time complexity of NHILS per generation is determined by three components:
	\begin{itemize}
		\item \textbf{Ranking and Selection:} Non-dominated sorting and crowding distance calculations require $O(S^2)$ for two objectives.
		\item \textbf{Local Search:} In the worst case, local search examines $O(mn)$ single-item swaps and $O(m^2 n^2)$ double-item swaps, where $m$ is the number of classes and $n$ is the average number of items per class. However, local search is applied only to a fraction $p_{LS}$ of the population, and in practice, it often terminates early due to the acceptance of improving moves.
		\item \textbf{Evaluation:} Using OPERA-MC, the evaluation cost is adaptive. While the worst-case cost per solution is $T_K$  (e.g., $10^6$) samples, early-stop mechanisms typically result in a much lower average cost.
	\end{itemize}
	
	Hence, the total complexity per generation is approximately $O(S^2 + p_{LS} S m^2 n^2 \cdot \bar{T})$, where $\bar{T}$ is the average number of samples used by OPERA-MC.
	This combination of hybrid initialization and item-wise local search empirically ensures robust convergence to high-fidelity Pareto approximations within practical time budgets.
	
	
	\section{Computational Study}
	\label{sec:experiments} 
	
	This section evaluates the performance of NHILS through a series of computational experiments.
	We first describe the benchmark instances and the compared algorithms.
	We then conduct a parameter sensitivity analysis to determine the appropriate configuration for local search.
	Finally, we compare NHILS against other algorithms using various performance metrics and provide an ablation study to validate the individual components of the algorithm.
	The source code and benchmark sets are available in an online repository\footnote{Github: \href{https://anonymous.4open.science/r/MO-CCMCKP-71EC/}{anonymous.4open.science/r/MO-CCMCKP-71EC/}}.
	All experiments were executed using Python 3.9.7 on a server equipped with an AMD EPYC 9754 processor (128 cores, 2.25 GHz), 512 GB RAM, and Ubuntu 22.04 LTS.

	\subsection{Benchmark Instances and Compared Algorithms}
	\label{subsec:benchmark_baselines}
	To comprehensively evaluate the performance of the proposed algorithm, we employ a rigorous experimental setup comprising diverse benchmark instanes and competitive baseline algorithms.
	
	\subsubsection{Benchmark Instances}
	
	We extended the benchmark sets from the single-objective CCMCKP study~\cite{Li2024chance} to a multi-objective context.
	The benchmark includes 12 instances with varying scales and characteristics, as summarized in Table~\ref{table:benchmark_parameters}.
	For all the instances, the chance constraint is fixed at $P_0 = 0.9$.
	
	\begin{table}[htpb]
		\centering
		\caption{Problem Instances setting for MO-CCMCKP.}
		\begin{tabular}{cccccc}
			\toprule
			\bfseries Instance & $m$ & $n$ & $L$ & $W^{LAB}$ & $W^{APP}$\\
			\midrule
			{\bfseries $ls1$} &10 &10 &500 &20 &35\\
			{\bfseries $ls2$} &10 &20 &500 &14 &15\\
			{\bfseries $ls3$} &20 &10 &500 &30 &41\\
			{\bfseries $ls4$} &30 &10 &500 &45 &60\\
			{\bfseries $ls5$} &40 &10 &500 &58 &87\\
			{\bfseries $ls6$} &50 &10 &500 &68 &97\\
			\bottomrule
		\end{tabular}
		\label{table:benchmark_parameters}
	\end{table}
	
	\begin{figure*}[htpb]
		\centering
		\subfloat[LAB]{\includegraphics[width=2.0\columnwidth]{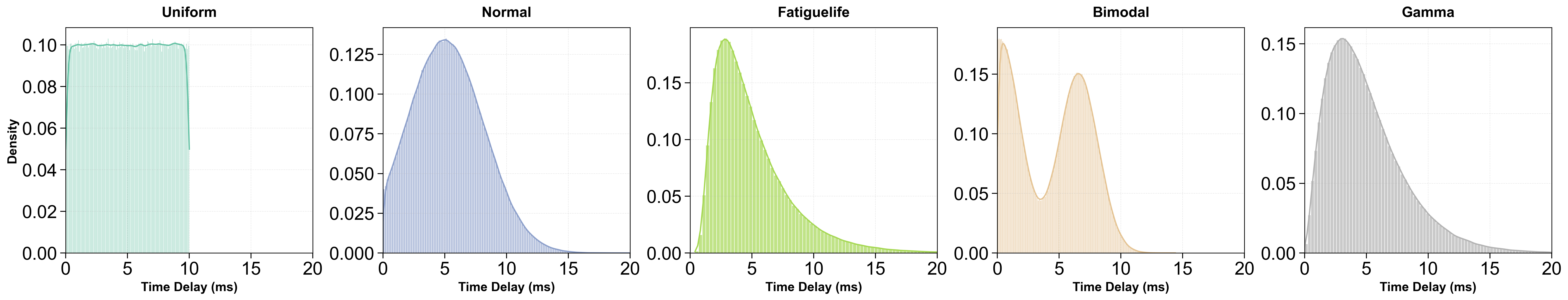}%
			\label{fig:pdf_LAB}}
		\hfil
		\subfloat[APP]{\includegraphics[width=2.0\columnwidth]{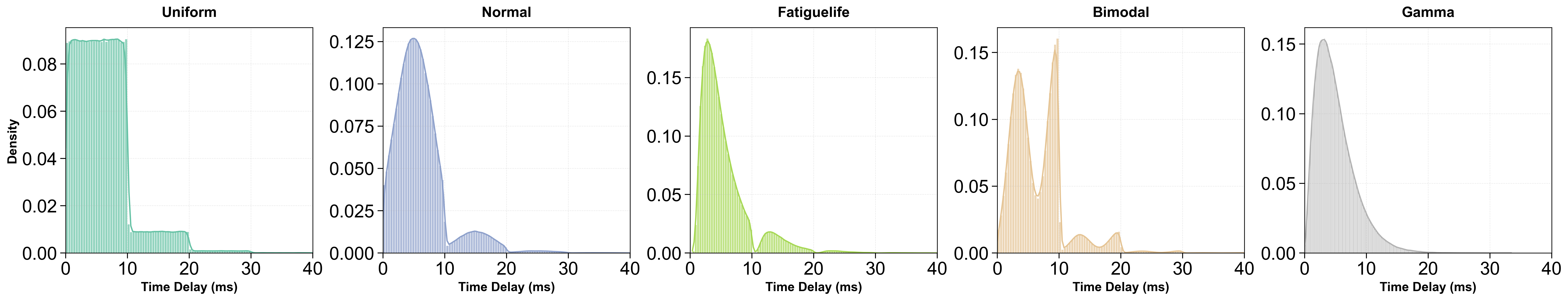}%
			\label{fig:pdf_APP}}
		\caption{Probability density function. The horizontal axis represents the time delay and the vertical axis represents the probability density.}
		\label{fig:APPLAB}
	\end{figure*}
	
	The \textbf{LAB} set consists of synthetic instances where item weights follow various continuous distributions, including uniform, truncated normal, fatigue-life, bimodal, and gamma distributions.
	
	The \textbf{APP} set comprises real-world instances simulating 5G network configurations from a Chinese telecommunications company, featuring a distinctive retransmission mechanism~\cite{nasrallah2018ultra,Li2024chance}. 
	As illustrated in Figure~\ref{fig:5G-multiobjective}, the end-to-end transmission path in a 5G network consists of multiple consecutive segments (e.g., radio access, backhaul, core network), each corresponding to a class in the CCMCKP. 
	For each segment, one implementation option (item) must be selected from several alternatives. 
	The total packet delay is the sum of delays from all selected options. 
	These delays are stochastic and lack closed-form distributions due to unpredictable factors such as workload and channel conditions.
	A retransmission mechanism is employed to enhance reliability: packets must be successfully transmitted within a fixed time window, with up to four attempts allowed. 
	As shown in Figure~\ref{fig:APPLAB}, the first attempt must complete within $(0,10]$ ms (success probability $0.9$). 
	Upon failure, a second attempt occurs within $(10,20]$ ms, also with success probability $0.9$, yielding a cumulative success probability of $0.09$. 
	The same logic applies to the third and fourth attempts, with delays accumulating. 
	If transmission fails after four attempts, the end-to-end transmission for that item is considered failed. 
	In this model, an item's ``weight'' corresponds to its stochastic transmission delay. 
	The network configuration problem seeks to select one option per segment to minimize total cost while ensuring that the probability of transmission success within four attempts meets a specified threshold.
	

	\subsubsection{Compared Algorithms}
	We evaluated NHILS against six representative multi-objective algorithms:
	(i) the widely-used NSGA-II \cite{Deb2002fast} and SPEA-II \cite{zitzler2001spea2} as foundational baselines;
	(ii) three variants of MOEA/D (using weighted sum, PBI, and Tchebycheff decomposition) that were proposed in~\cite{hewa2024using} for dynamic chance-constrained knapsack problem and could be adapted to our problem; and
	(iii) the recently proposed ANSGA-II~\cite{chen2025altruistic} as a more state-of-the-art variant of NSGA-II. 
	All algorithms were implemented using the Pymoo framework with identical solution evaluation approach to ensure fair comparison.
	For ANSGA-II, we implemented the algorithm following the original paper specifications, including the altruistic nurturing process, adaptive abandonment threshold, and double selection strategy. 
	All parameters were set to their recommended values from the literature: $\beta=0.4$, $\delta=0.8$, and $k=4$ for the nearest neighbor selection. 
	
	\subsection{Experimental Settings and Sensitivity Analysis}
	\label{subsec:experimental_settings}
	
	\subsubsection{Performance Metrics}
	We use four metrics to evaluate the algorithms:
	\begin{itemize}
		\item \textbf{Hypervolume (HV)}: Measures the volume of the objective space dominated by the obtained Pareto front relative to a reference point. 
		Higher HV values indicate better convergence and diversity.
		The reference point is set to be slightly worse than the nadir point constructed from all non-dominated solutions found by all algorithms in all 30 experimental runs.
		\item \textbf{Inverted Generational Distance (IGD)}: Computes the average distance from the points in the true Pareto front to the nearest solution in the obtained front. 
		Lower IGD values reflect better convergence to the true Pareto front. 
		\item \textbf{IGD+}: An enhanced version of IGD that satisfies Pareto compliance properties, providing a more reliable assessment of solution quality.
		For the calculation of both IGD and IGD+, we approximate the true Pareto front by constructing a reference set. This set is formed by merging the non-dominated solutions from all algorithms across all independent runs and applying a filtering process to retain only the global non-dominated solutions.
		\item \textbf{Feasible Solution Ratio (FSR)}: Measured by the proportion of solutions in the final population that meet the confidence constraint, based on their real confidence level (RCL). The RCL is estimated from $10^7$ simulated samples per solution.
		\item \textbf{Generations (Gens)}: The generations that populations iterate until termination.
	\end{itemize}
	
	\subsubsection{Termination Criterion}
	To ensure fair comparison, we adopt a time-based termination criterion.
	It is motivated by the fact that the computational cost is overwhelmingly dominated by solution evaluation (involving MC simulations) which is time-consuming.
	NHILS is run for 100 generations, and its execution time is recorded for each instance.
	Compared algorithms are allowed to run until they reach this recorded time.
	This protocol ensures that all algorithms consume equivalent computational resources.
	Results are averaged over 30 independent runs on each problem instance.
	
	\subsubsection{Parameter Settings}
	Part of the parameters of NHILS is set following the DDALS for solving single-objective CCMCKP~\cite{Li2024chance}.
	The surrogate weight parameter is set to $\lambda = 3$, which effectively balances the consideration of mean and standard deviation in the weight estimation.
	The population size is $N = 100$.
	We apply simulated binary crossover with probability $p_c = 0.9$ and distribution index $\eta_c = 15$, and polynomial mutation with probability $p_m = 1/m$ and distribution index $\eta_m = 20$.
	
	To determine the OPERA-MC parameters, we conducted preliminary tests on the LAB-ls2 and APP-ls2 instances.
	Table~\ref{table:mc_empirical_lab} and Table~\ref{table:mc_empirical_app} present the absolute errors between estimated and true CLs for different sample counts across various $W$ values.
	Since a portion of the solutions in the real-world scenario we consider exhibit very high confidence levels ($>0.999$), for a target $P_0 = 0.9$, we configure the algorithms with the following parameter values:
	\begin{align*}
		T &= [10^4, 10^5, 10^6] \quad \text{(cumulative sample sizes)} \\
		C &= [0.999, 0.9999, \infty] \quad \text{(confidence thresholds)}
	\end{align*}
	
	This configuration derived from extensive simulations shows that:
	\begin{itemize}
		\item Solutions with confidence $p \leq 0.999$ can be reliably distinguished with $10^4$ samples.
		\item Solutions in the range $0.999 < p \leq 0.9999$ require about $10^5$ samples for accurate estimation.
		\item Solutions with $p > 0.9999$ (near the Pareto front) need the full $10^6$ samples for precise comparison.
	\end{itemize}
	
	The adaptive nature of OPERA-MC makes it robust to variations in problem characteristics. 
	Even if the chosen thresholds are not optimal, it still provides significant speedups while maintaining correct dominance relationships with high probability.
	
	\begin{table}[htpb]
		\caption{Simulation Count and Estimation Error on LAB-ls2}
		\label{table:mc_empirical_lab}
		\centering
		\begin{tabular}{c|ccccc}
			\hline
			\multirow{2}{*}{ST} & \multicolumn{5}{c}{Absolute Error for Different $W$ Values} \\
			\cline{2-6}
			& 13.2 & 14.7 & 15.9 & 16.9 & 17.9 \\
			\hline
			$10^2$ & 0.013997 & 0.007000 & 0.001702 & 8.74E-05 & 7.10E-06 \\
			$10^3$ & 0.008203 & 0.003251 & 0.000349 & 8.74E-05 & 7.10E-06 \\
			$10^4$ & 0.009426 & 0.003095 & 0.000448 & 4.76E-05 & 1.57E-05 \\
			$10^5$ & 0.008073 & 0.000737 & 6.59E-05 & 2.00E-05 & 1.43E-05 \\
			$10^6$ & 0.007041 & 0.000975 & 0.000427 & 3.85E-05 & 2.90E-06 \\
			\hline
		\end{tabular}
	\end{table}
	
	\begin{table}[htpb]
		\caption{Simulation Count and Estimation Error on APP-ls2}
		\label{table:mc_empirical_app}
		\centering
		\begin{tabular}{c|ccccc}
			\hline
			\multirow{2}{*}{ST} & \multicolumn{5}{c}{Absolute Error for Different $W$ Values} \\
			\cline{2-6}
			& 13.2 & 14.6 & 15.8 & 16.8 & 17.8 \\
			\hline
			$10^2$ & 0.025311 & 0.008009 & 0.000953 & 9.58E-05 & 7.90E-06 \\
			$10^3$ & 0.006966 & 0.002918 & 0.000700 & 9.58E-05 & 7.90E-06 \\
			$10^4$ & 0.017886 & 0.003295 & 0.000262 & 7.34E-05 & 7.90E-06 \\
			$10^5$ & 0.004049 & 0.001846 & 0.000439 & 3.02E-05 & 8.68E-06 \\
			$10^6$ & 0.000992 & 0.000829 & 0.000248 & 3.73E-05 & 3.52E-06 \\
			\hline
		\end{tabular}
	\end{table}
	
	\begin{table}[htpb]
		\caption{Sensitivity Analysis of Local Search Operator Calling Probability $p_{LS}$ for NHILS}
		\label{table:parameter_sensitivity}
		\centering
		\begin{tabular}{c|ccc}
			\hline
			\textbf{$p_{LS}$} & \textbf{HV} & \textbf{IGD} & \textbf{Runtime(s)} \\
			\hline
			0.02 & 5.73 $\pm$ 0.15 & 0.55 $\pm$ 0.08 & 268.9 $\pm$ 24.3 \\
			0.05 & 5.79 $\pm$ 0.12 & 0.40 $\pm$ 0.07 & 490.1 $\pm$ 32.7 \\
			0.10 & 5.83 $\pm$ 0.09 & 0.29 $\pm$ 0.05 & 859.3 $\pm$ 45.8 \\
			0.20 & 5.84 $\pm$ 0.08 & 0.28 $\pm$ 0.04 & 1664.7 $\pm$ 68.9 \\
			0.30 & 5.85 $\pm$ 0.07 & 0.25 $\pm$ 0.03 & 2440.7 $\pm$ 92.1 \\
			\hline
		\end{tabular}
	\end{table}
	
	\subsubsection{Sensitivity of Local Search Probability ($p_{LS}$)}
	We conduct a sensitivity analysis to determine the calling probability $p_{LS}$ for local search. 
	The analysis evaluates five probability values: $p_{LS} \in \{0.02, 0.05, 0.1, 0.2, 0.3\}$ across all benchmark instances.
	Table~\ref{table:parameter_sensitivity} presents the aggregated results. 
	Low probabilities ($p_{LS} = 0.02, 0.05$) result in shorter runtimes but suboptimal solution quality, as indicated by lower HV and higher IGD values.
	High probabilities ($p_{LS} = 0.2, 0.3$) yield marginal improvements in solution quality but incur substantial computational overhead, with runtimes increasing by 94\% and 184\% respectively compared to $p_{LS} = 0.1$.
	The probability $p_{LS} = 0.1$ provides an optimal balance, achieving near-maximal solution quality (HV = 5.83, IGD = 0.29) with reasonable computational cost (859.32 seconds).
	
	\subsection{Results and Analysis}
	\label{subsec:main_results}
	We compared NHILS with the baselinses on both LAB and APP instances, providing insights into algorithm performance across diverse problem characteristics.

	\begin{table*}[htbp]
		\centering
		\caption{Performance Comparison of Algorithms Across LAB Instances}
		\label{tab:complete_performance_lab}
		\resizebox{\textwidth}{!}{%
			\begin{tabular}{llcccccccc}
				\toprule
				\textbf{Test Case} & \textbf{Indicators} & \textbf{NHILS} & \textbf{NSGA-II} & \textbf{SPEA-II} & \textbf{ANSGA-II} & \textbf{MOEA/D(ws)} & \textbf{MOEA/D(pbi)} & \textbf{MOEA/D(tche)} \\
				\midrule
				
				\multirow{5}{*}{LAB-ls1}
				& HV & $\mathbf{0.643 \pm 0.153}$ & $0.546 \pm 0.163$ & $0.545 \pm 0.079$ & $0.565 \pm 0.110$ & $0.241 \pm 0.272$ & $0.243 \pm 0.247$ & $0.238 \pm 0.239$ \\
				& IGD & $0.322 \pm 0.097$ & $0.320 \pm 0.107$ & $\mathbf{0.315 \pm 0.106}$ & $0.339 \pm 0.097$ & $\infty$ & $\infty$ & $\infty$ \\
				& IGD+ & $\mathbf{0.230 \pm 0.146}$ & $0.271 \pm 0.127$ & $0.271 \pm 0.111$ & $0.286 \pm 0.117$ & $\infty$ & $\infty$ & $\infty$ \\
				& FSR & $\mathbf{1.000 \pm 0.000}$ & $0.992 \pm 0.027$ & $0.997 \pm 0.015$ & $\mathbf{1.000 \pm 0.000}$ & $0.940 \pm 0.185$ & $0.900 \pm 0.305$ & $0.927 \pm 0.252$ \\
				& \#Gens & $100$ & $1231 \pm 974$ & $1507 \pm 1076$ & $690 \pm 563$ & $9607 \pm 3994$ & $8275 \pm 3442$ & $9416 \pm 3919$ \\
				\midrule
				
				\multirow{5}{*}{LAB-ls2}
				& HV & $2.958 \pm 0.053$ & $3.176 \pm 0.692$ & $3.275 \pm 0.631$ & $\mathbf{3.386 \pm 0.798}$ & $2.485 \pm 0.720$ & $2.031 \pm 0.790$ & $2.707 \pm 0.795$ \\
				& IGD & $0.394 \pm 0.009$ & $0.405 \pm 0.080$ & $0.385 \pm 0.066$ & $\mathbf{0.366 \pm 0.090}$ & $0.508 \pm 0.109$ & $\infty$ & $0.472 \pm 0.100$ \\
				& IGD+ & $0.354 \pm 0.015$ & $0.333 \pm 0.115$ & $0.312 \pm 0.105$ & $\mathbf{0.286 \pm 0.123}$ & $0.455 \pm 0.131$ & $\infty$ & $0.409 \pm 0.130$ \\
				& FSR & $\mathbf{1.000 \pm 0.000}$ & $0.989 \pm 0.020$ & $\mathbf{0.996 \pm 0.012}$ & $0.988 \pm 0.028$ & $0.988 \pm 0.035$ & $0.962 \pm 0.183$ & $0.979 \pm 0.036$ \\
				& \#Gens & $100$ & $1733 \pm 1914$ & $1944 \pm 2549$ & $735 \pm 754$ & $6713 \pm 2279$ & $5808 \pm 1976$ & $6596 \pm 2225$ \\
				\midrule
				
				\multirow{5}{*}{LAB-ls3}
				& HV & $\mathbf{0.428 \pm 0.050}$ & $0.073 \pm 0.115$ & $0.054 \pm 0.093$ & $0.166 \pm 0.170$ & $0.002 \pm 0.011$ & $0.017 \pm 0.050$ & $0.033 \pm 0.101$ \\
				& IGD & $\mathbf{0.050 \pm 0.035}$ & $0.203 \pm 0.108$ & $0.238 \pm 0.128$ & $0.153 \pm 0.133$ & $\infty$ & $\infty$ & $\infty$ \\
				& IGD+ & $\mathbf{0.049 \pm 0.036}$ & $0.203 \pm 0.108$ & $0.238 \pm 0.128$ & $0.152 \pm 0.133$ & $\infty$ & $\infty$ & $\infty$ \\
				& FSR & $\mathbf{0.988 \pm 0.012}$ & $\mathbf{0.988 \pm 0.013}$ & $\mathbf{0.989 \pm 0.013}$ & $0.983 \pm 0.015$ & $0.892 \pm 0.252$ & $0.885 \pm 0.197$ & $0.869 \pm 0.297$ \\
				& \#Gens & $100$ & $2739 \pm 2454$ & $1232 \pm 1653$ & $1018 \pm 1175$ & $9553 \pm 557$ & $8311 \pm 463$ & $9377 \pm 558$ \\
				\midrule
				
				\multirow{5}{*}{LAB-ls4}
				& HV & $\mathbf{0.706 \pm 0.036}$ & $0.009 \pm 0.050$ & $0.000 \pm 0.002$ & $0.022 \pm 0.066$ & $0.020 \pm 0.063$ & $0.015 \pm 0.041$ & $0.024 \pm 0.069$ \\
				& IGD & $\mathbf{0.041 \pm 0.022}$ & $0.407 \pm 0.089$ & $0.444 \pm 0.084$ & $0.402 \pm 0.101$ & $0.430 \pm 0.113$ & $0.509 \pm 0.135$ & $0.396 \pm 0.142$ \\
				& IGD+ & $\mathbf{0.037 \pm 0.025}$ & $0.407 \pm 0.090$ & $0.443 \pm 0.084$ & $0.402 \pm 0.102$ & $0.430 \pm 0.113$ & $0.509 \pm 0.135$ & $0.396 \pm 0.142$ \\
				& FSR & $\mathbf{0.989 \pm 0.020}$ & $\mathbf{0.991 \pm 0.015}$ & $\mathbf{0.990 \pm 0.016}$ & $\mathbf{0.993 \pm 0.012}$ & $0.988 \pm 0.026$ & $0.971 \pm 0.050$ & $\mathbf{0.990 \pm 0.018}$ \\
				& \#Gens & $100$ & $2372 \pm 1009$ & $2711 \pm 1869$ & $1470 \pm 950$ & $34645 \pm 2613$ & $29636 \pm 2207$ & $33857 \pm 2433$ \\
				\midrule
				
				\multirow{5}{*}{LAB-ls5}
				& HV & $\mathbf{4.647 \pm 0.265}$ & - & - & - & $0.312 \pm 0.538$ & $0.284 \pm 0.665$ & $0.572 \pm 0.902$ \\
				& IGD & $\mathbf{0.196 \pm 0.039}$ & - & - & - & $\infty$ & $\infty$ & $\infty$ \\
				& IGD+ & $\mathbf{0.067 \pm 0.043}$ & - & - & - & $\infty$ & $\infty$ & $\infty$ \\
				& FSR & $\mathbf{0.995 \pm 0.009}$ & - & - & - & $0.516 \pm 0.465$ & $0.374 \pm 0.447$ & $0.462 \pm 0.480$ \\
				& \#Gens & $\mathbf{100 \pm 0}$ & - & - & - & $4861 \pm 337$ & $4233 \pm 289$ & $4789 \pm 337$ \\
				\midrule
				
				\multirow{5}{*}{LAB-ls6}
				& HV & $\mathbf{1.668 \pm 0.442}$ & - & - & - & $0.000 \pm 0.000$ & $0.000 \pm 0.000$ & $0.000 \pm 0.000$ \\
				& IGD & $\mathbf{0.123 \pm 0.101}$ & - & - & - & $\infty$ & $\infty$ & $\infty$ \\
				& IGD+ & $\mathbf{0.115 \pm 0.106}$ & - & - & - & $\infty$ & $\infty$ & $\infty$ \\
				& FSR & $\mathbf{0.997 \pm 0.007}$ & - & - & - & $0.333 \pm 0.451$ & $0.200 \pm 0.377$ & $0.376 \pm 0.447$ \\
				& \#Gens & $100$ & - & - & - & $14109 \pm 5156$ & $12333 \pm 4500$ & $13899 \pm 5087$ \\
				\midrule
				
				\multicolumn{2}{l}{\textbf{NHILS vs Baseline (Win/Draw/Lose)}} \\
				\multicolumn{2}{l}{HV} & - & 5/0/1 & 5/0/1 & 5/0/1 & 6/0/0 & 6/0/0 & 6/0/0 \\
				\multicolumn{2}{l}{IGD} & - & 5/0/1 & 4/0/2 & 5/0/1 & 6/0/0 & 6/0/0 & 6/0/0 \\
				\multicolumn{2}{l}{IGD+} & - & 5/0/1 & 5/0/1 & 5/0/1 & 6/0/0 & 6/0/0 & 6/0/0 \\
				\multicolumn{2}{l}{FSR} & - & 4/2/0 & 3/3/0 & 3/3/0 & 6/0/0 & 6/0/0 & 5/1/0 \\
				\bottomrule
			\end{tabular}%
		}
	\end{table*}
	
	\subsubsection{Comparison on LAB and APP Instances}
	In Tables \ref{tab:complete_performance_lab} and \ref{tab:complete_performance_app}, all presented results are derived from a statistical analysis conducted with 30 repetitions, denoted as ``mean standard $\pm$ deviation''.
	For each performance metric (HV, IGD, IGD+, and FSR), we apply Friedman tests with post-hoc Wilcoxon signed-rank tests at a 5\% significance level to identify algorithms that are not statistically inferior to any competitor. 
	Bold values indicate algorithms that either achieve the best performance or show no significant difference from the best-performing algorithm for each metric-instance combination.
	``mean $\pm$ standard deviation'' of performance metrics are reported to account for stochastic variations and ensure statistical CL.
	``$\infty$'' indicates that the algorithm failed to converge to the feasible region, resulting in infinite distance metrics.
	``-'' denotes complete algorithm failure, where no feasible solutions were obtained across all 30 independent runs.
	Additionally, we conduct pairwise comparisons between NHILS and each baseline algorithm, reporting Win/Draw/Lose counts across all test instances. 
	The ``Win/Draw/Lose'' analysis at the bottom of each table summarizes NHILS's comparative performance against each baseline algorithm across all six instances for each metric. 
	A ``Win'' indicates NHILS demonstrates statistically superior performance, a ``Draw'' signifies no statistically significant difference, and a ``Lose'' denotes NHILS is statistically inferior.
	
	Table~\ref{tab:complete_performance_lab} demonstrates NHILS's strong performance on LAB instances, particularly in larger-scale scenarios. 
	NHILS achieves statistically superior performance in 22 out of 30 metric-instance combinations (73.3\%), showcasing significant advantages in convergence quality and computational efficiency. 
	In LAB-ls3 and LAB-ls4, NHILS substantially outperforms other algorithms across all metrics.
	For instance, in LAB-ls4, NHILS attains HV = $0.706 \pm 0.036$ compared to the second-best ANSGA-II (HV = $0.022 \pm 0.066$), indicating superior convergence and diversity. 
	NHILS consistently maintains high feasibility satisfaction ratios (FSR $\geq$ 0.988) across all LAB instances, while MOEA/D variants exhibit catastrophic failures in maintaining feasibility in larger problems (e.g., FSR $\approx$ 0.20--0.52 in LAB-ls5 and LAB-ls6). 
	The ``Win/Draw/Lose'' analysis confirms NHILS's dominance with perfect 6/0/0 records against MOEA/D variants in most metrics, and competitive performance against NSGA-II, SPEA-II, and ANSGA-II.
	
	\begin{table*}[htbp]
		\centering
		\caption{Performance Comparison of Algorithms Across APP Instances}
		\label{tab:complete_performance_app}
		\resizebox{\textwidth}{!}{%
			\begin{tabular}{llcccccccc}
				\toprule
				\textbf{Test Case} & \textbf{Indicators} & \textbf{NHILS} & \textbf{NSGA-II} & \textbf{SPEA-II} & \textbf{ANSGA-II} & \textbf{MOEA/D(ws)} & \textbf{MOEA/D(pbi)} & \textbf{MOEA/D(tche)} \\
				\midrule
				
				\multirow{5}{*}{APP-ls1}
				& HV & $\mathbf{0.368 \pm 0.027}$ & $0.320 \pm 0.084$ & $0.321 \pm 0.067$ & $0.331 \pm 0.062$ & $0.202 \pm 0.125$ & $0.200 \pm 0.073$ & $0.244 \pm 0.073$ \\
				& IGD & $0.039 \pm 0.039$ & $0.045 \pm 0.027$ & $0.042 \pm 0.025$ & $\mathbf{0.035 \pm 0.019}$ & $0.084 \pm 0.068$ & $0.131 \pm 0.059$ & $0.063 \pm 0.060$ \\
				& IGD+ & $0.035 \pm 0.042$ & $0.044 \pm 0.028$ & $0.041 \pm 0.026$ & $\mathbf{0.032 \pm 0.022}$ & $0.082 \pm 0.070$ & $0.131 \pm 0.059$ & $0.061 \pm 0.060$ \\
				& FSR & $\mathbf{0.999 \pm 0.005}$ & $0.988 \pm 0.011$ & $0.992 \pm 0.007$ & $0.988 \pm 0.010$ & $0.943 \pm 0.039$ & $0.913 \pm 0.032$ & $0.941 \pm 0.032$ \\
				& \#Gens & $100$ & $1340 \pm 1055$ & $1129 \pm 876$ & $775 \pm 724$ & $8953 \pm 4124$ & $7729 \pm 3569$ & $8763 \pm 4051$ \\
				\midrule
				
				\multirow{5}{*}{APP-ls2}
				& HV & $\mathbf{0.582 \pm 0.061}$ & $0.338 \pm 0.207$ & $0.303 \pm 0.190$ & $0.394 \pm 0.172$ & $0.228 \pm 0.173$ & $0.189 \pm 0.160$ & $0.264 \pm 0.203$ \\
				& IGD & $\mathbf{0.050 \pm 0.054}$ & $0.233 \pm 0.262$ & $0.302 \pm 0.297$ & $0.167 \pm 0.183$ & $\infty$ & $\infty$ & $\infty$ \\
				& IGD+ & $\mathbf{0.046 \pm 0.056}$ & $0.233 \pm 0.263$ & $0.301 \pm 0.297$ & $0.166 \pm 0.184$ & $\infty$ & $\infty$ & $\infty$ \\
				& FSR & $\mathbf{0.994 \pm 0.006}$ & $0.991 \pm 0.012$ & $\mathbf{0.996 \pm 0.011}$ & $0.989 \pm 0.011$ & $0.802 \pm 0.371$ & $0.857 \pm 0.343$ & $0.855 \pm 0.342$ \\
				& \#Gens & $100$ & $1639 \pm 2908$ & $2435 \pm 2745$ & $1311 \pm 1420$ & $8427 \pm 3979$ & $7281 \pm 3427$ & $8234 \pm 3880$ \\
				\midrule
				
				\multirow{5}{*}{APP-ls3}
				& HV & $\mathbf{1.121 \pm 0.063}$ & $0.922 \pm 0.176$ & $0.877 \pm 0.190$ & $0.915 \pm 0.137$ & $0.409 \pm 0.288$ & $0.671 \pm 0.207$ & $0.767 \pm 0.228$ \\
				& IGD & $\mathbf{0.055 \pm 0.029}$ & $0.148 \pm 0.076$ & $0.170 \pm 0.080$ & $0.159 \pm 0.068$ & $0.486 \pm 0.226$ & $0.284 \pm 0.117$ & $\infty$ \\
				& IGD+ & $\mathbf{0.043 \pm 0.035}$ & $0.145 \pm 0.079$ & $0.168 \pm 0.082$ & $0.157 \pm 0.070$ & $0.486 \pm 0.226$ & $0.284 \pm 0.117$ & $\infty$ \\
				& FSR & $0.957 \pm 0.032$ & $\mathbf{0.965 \pm 0.034}$ & $0.953 \pm 0.037$ & $\mathbf{0.965 \pm 0.036}$ & $0.915 \pm 0.036$ & $0.911 \pm 0.031$ & $0.894 \pm 0.171$ \\
				& \#Gens & $100$ & $1047 \pm 503$ & $1470 \pm 630$ & $624 \pm 545$ & $14938 \pm 1885$ & $12751 \pm 1665$ & $14668 \pm 1988$ \\
				\midrule
				
				\multirow{5}{*}{APP-ls4}
				& HV & $\mathbf{2.796 \pm 0.086}$ & $1.849 \pm 0.484$ & $1.733 \pm 0.430$ & $1.690 \pm 0.449$ & $0.924 \pm 0.584$ & $1.397 \pm 0.441$ & $1.327 \pm 0.463$ \\
				& IGD & $\mathbf{0.105 \pm 0.012}$ & $0.176 \pm 0.053$ & $0.178 \pm 0.055$ & $0.181 \pm 0.056$ & $0.318 \pm 0.150$ & $0.258 \pm 0.106$ & $0.195 \pm 0.072$ \\
				& IGD+ & $\mathbf{0.043 \pm 0.021}$ & $0.157 \pm 0.065$ & $0.167 \pm 0.062$ & $0.170 \pm 0.061$ & $0.317 \pm 0.151$ & $0.254 \pm 0.110$ & $0.193 \pm 0.074$ \\
				& FSR & $\mathbf{0.997 \pm 0.005}$ & $0.957 \pm 0.030$ & $0.962 \pm 0.025$ & $0.959 \pm 0.031$ & $0.894 \pm 0.080$ & $0.889 \pm 0.057$ & $0.920 \pm 0.044$ \\
				& \#Gens & $100$ & $10018 \pm 6025$ & $9531 \pm 6296$ & $4488 \pm 3198$ & $29773 \pm 2195$ & $25788 \pm 1801$ & $29071 \pm 2245$ \\
				\midrule
				
				\multirow{5}{*}{APP-ls5}
				& HV & $\mathbf{2.334 \pm 0.112}$ & $1.195 \pm 0.408$ & $1.046 \pm 0.430$ & $1.378 \pm 0.385$ & $1.186 \pm 0.412$ & $1.054 \pm 0.506$ & $1.232 \pm 0.390$ \\
				& IGD & $\mathbf{0.066 \pm 0.031}$ & $0.221 \pm 0.066$ & $0.247 \pm 0.102$ & $0.186 \pm 0.065$ & $0.195 \pm 0.075$ & $\infty$ & $0.190 \pm 0.073$ \\
				& IGD+ & $\mathbf{0.040 \pm 0.035}$ & $0.219 \pm 0.071$ & $0.247 \pm 0.102$ & $0.184 \pm 0.068$ & $0.195 \pm 0.075$ & $\infty$ & $0.189 \pm 0.074$ \\
				& FSR & $\mathbf{0.993 \pm 0.011}$ & $0.968 \pm 0.017$ & $0.966 \pm 0.019$ & $0.973 \pm 0.019$ & $0.913 \pm 0.034$ & $0.830 \pm 0.166$ & $0.876 \pm 0.057$ \\
				& \#Gens & $100$ & $2304 \pm 77$ & $2272 \pm 68$ & $1165 \pm 26$ & $5535 \pm 127$ & $4835 \pm 114$ & $5433 \pm 122$ \\
				\midrule
				
				\multirow{5}{*}{APP-ls6}
				& HV & $\mathbf{4.959 \pm 0.295}$ & - & - & - & $0.721 \pm 1.227$ & $0.920 \pm 1.534$ & $1.358 \pm 1.753$ \\
				& IGD & $\mathbf{0.131 \pm 0.044}$ & - & - & - & $\infty$ & $\infty$ & $\infty$ \\
				& IGD+ & $\mathbf{0.059 \pm 0.051}$ & - & - & - & $\infty$ & $\infty$ & $\infty$ \\
				& FSR & $\mathbf{0.985 \pm 0.016}$ & - & - & - & $0.252 \pm 0.316$ & $0.162 \pm 0.275$ & $0.296 \pm 0.346$ \\
				& \#Gens & $100$ & - & - & - & $7815 \pm 200$ & $6826 \pm 167$ & $7684 \pm 201$ \\
				\midrule
				
				\multicolumn{2}{l}{\textbf{NHILS vs Baseline (Win/Draw/Lose)}} \\
				\multicolumn{2}{l}{HV} & - & 6/0/0 & 6/0/0 & 6/0/0 & 6/0/0 & 6/0/0 & 6/0/0 \\
				\multicolumn{2}{l}{IGD} & - & 6/0/0 & 6/0/0 & 5/0/1 & 6/0/0 & 6/0/0 & 6/0/0 \\
				\multicolumn{2}{l}{IGD+} & - & 6/0/0 & 6/0/0 & 5/0/1 & 6/0/0 & 6/0/0 & 6/0/0 \\
				\multicolumn{2}{l}{FSR} & - & 5/0/1 & 5/1/0 & 5/0/1 & 6/0/0 & 6/0/0 & 6/0/0 \\
				\bottomrule
			\end{tabular}%
		}
	\end{table*}
	
	As shown in Table \ref{tab:complete_performance_app}, NHILS exhibits statistically superior performance in 23 out of 30 metric-instance combinations (76.7\%) on APP instances.
	The algorithm demonstrates remarkable consistency and scalability, with HV values increasing from $0.368 \pm 0.027$ in APP-ls1 to $4.959 \pm 0.295$ in APP-ls6, reflecting its ability to handle increasing problem complexity. 
	In APP-ls4, NHILS achieves an IGD+ value of $0.043 \pm 0.021$, significantly better than NSGA-II ($0.157 \pm 0.065$), SPEA-II ($0.167 \pm 0.062$), and ANSGA-II ($0.170 \pm 0.061$), indicating more accurate approximation of the Pareto front. 
	NHILS maintains consistently high feasibility (FSR $\geq$ 0.957) across all APP instances, while competitors show degraded performance in larger problems. 
	The comprehensive 6/0/0 ``Win/Draw/Lose'' records against MOEA/D variants across all metrics underscore NHILS's superiority in convergence, diversity, and feasibility maintenance.
	
	\begin{figure*}[htpb]
		\centering
		
		\subfloat[LAB-ls1]{\includegraphics[width=0.6\columnwidth]{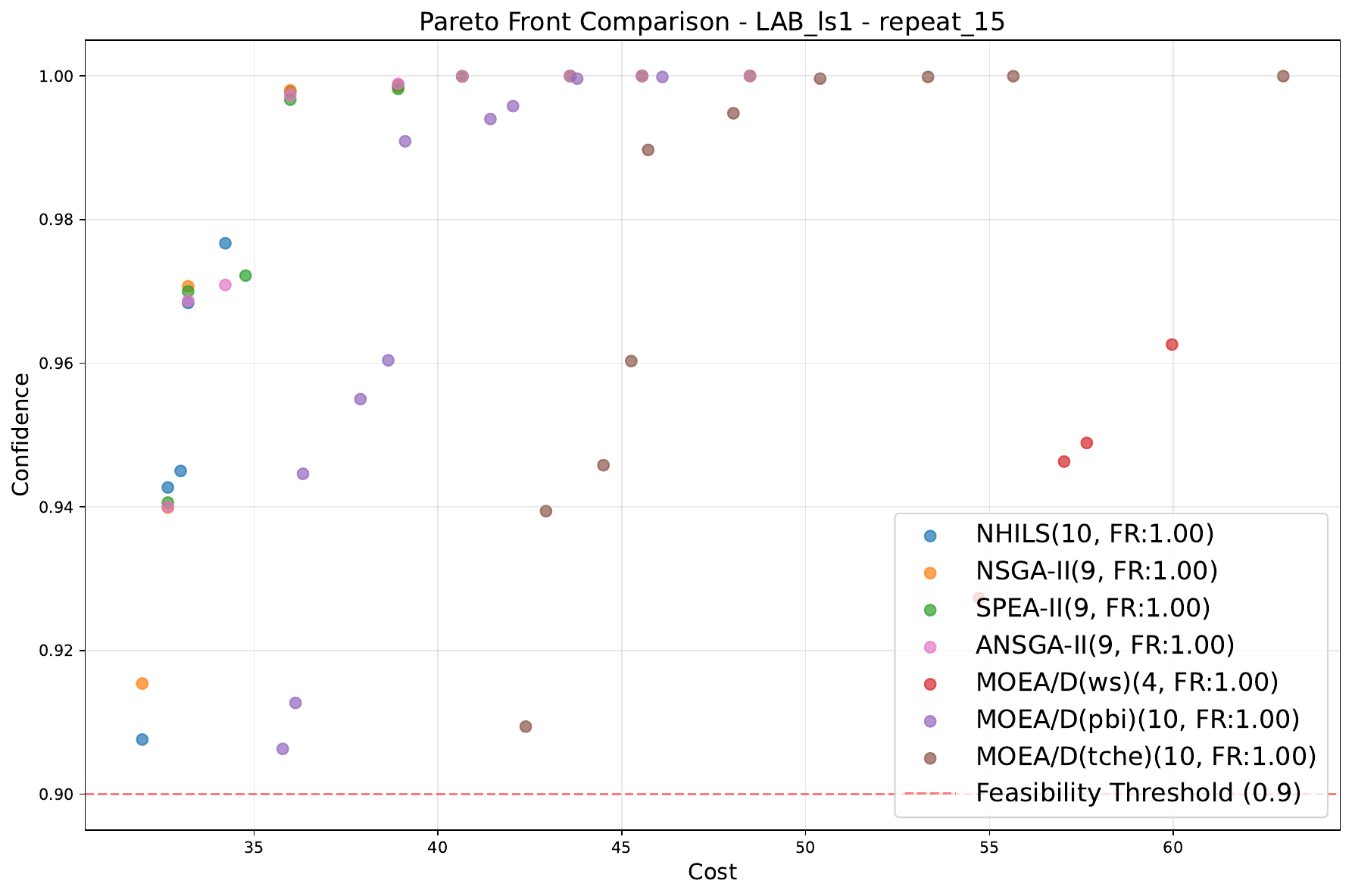}%
			\label{fig:LAB-ls1}}
		\hfil
		\subfloat[LAB-ls2]{\includegraphics[width=0.6\columnwidth]{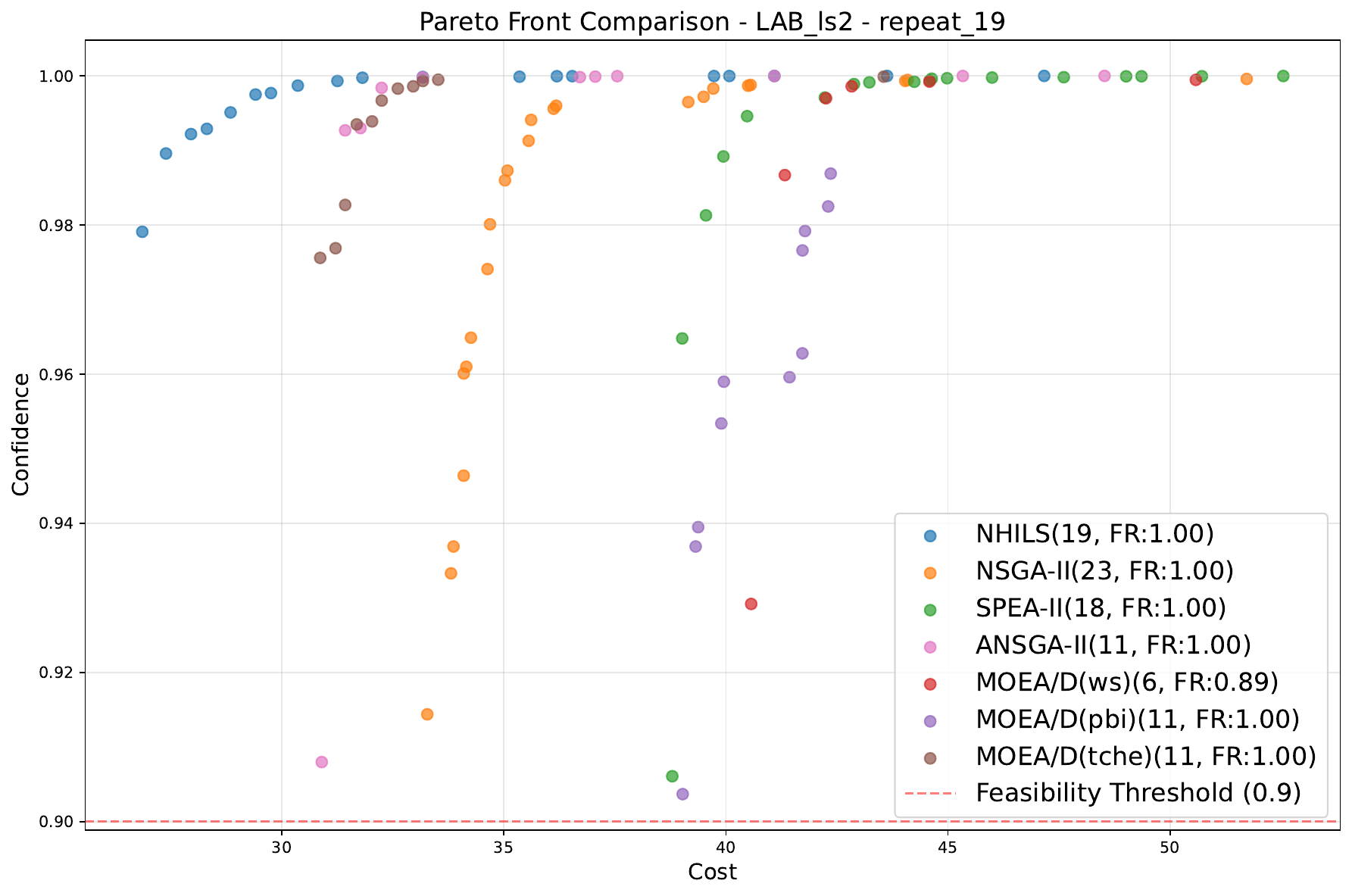}%
			\label{fig:LAB-ls2}}
		\hfil
		\subfloat[LAB-ls3]{\includegraphics[width=0.6\columnwidth]{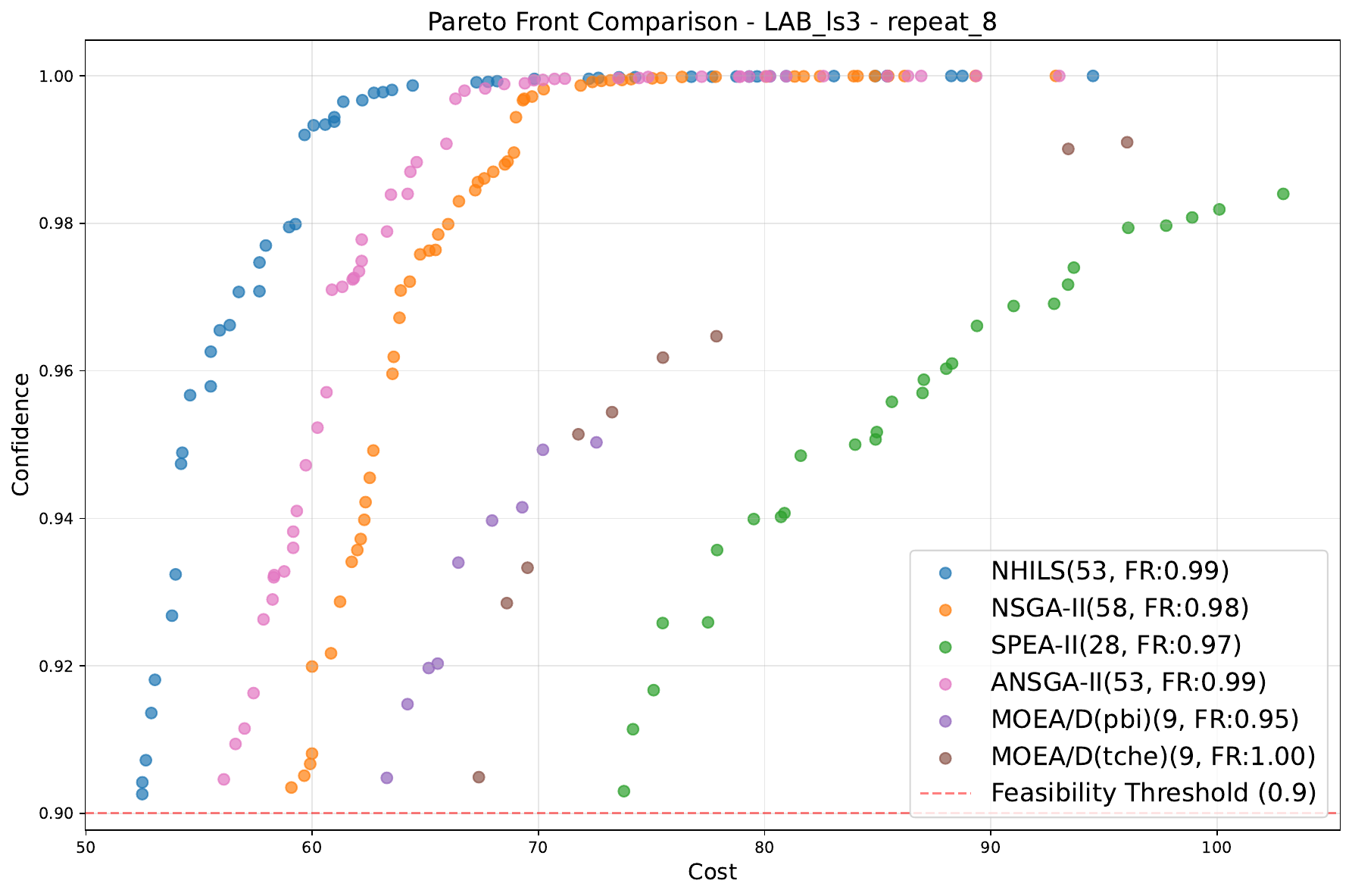}%
			\label{fig:LAB-ls3}}
		\hfil
		\subfloat[LAB-ls4]{\includegraphics[width=0.6\columnwidth]{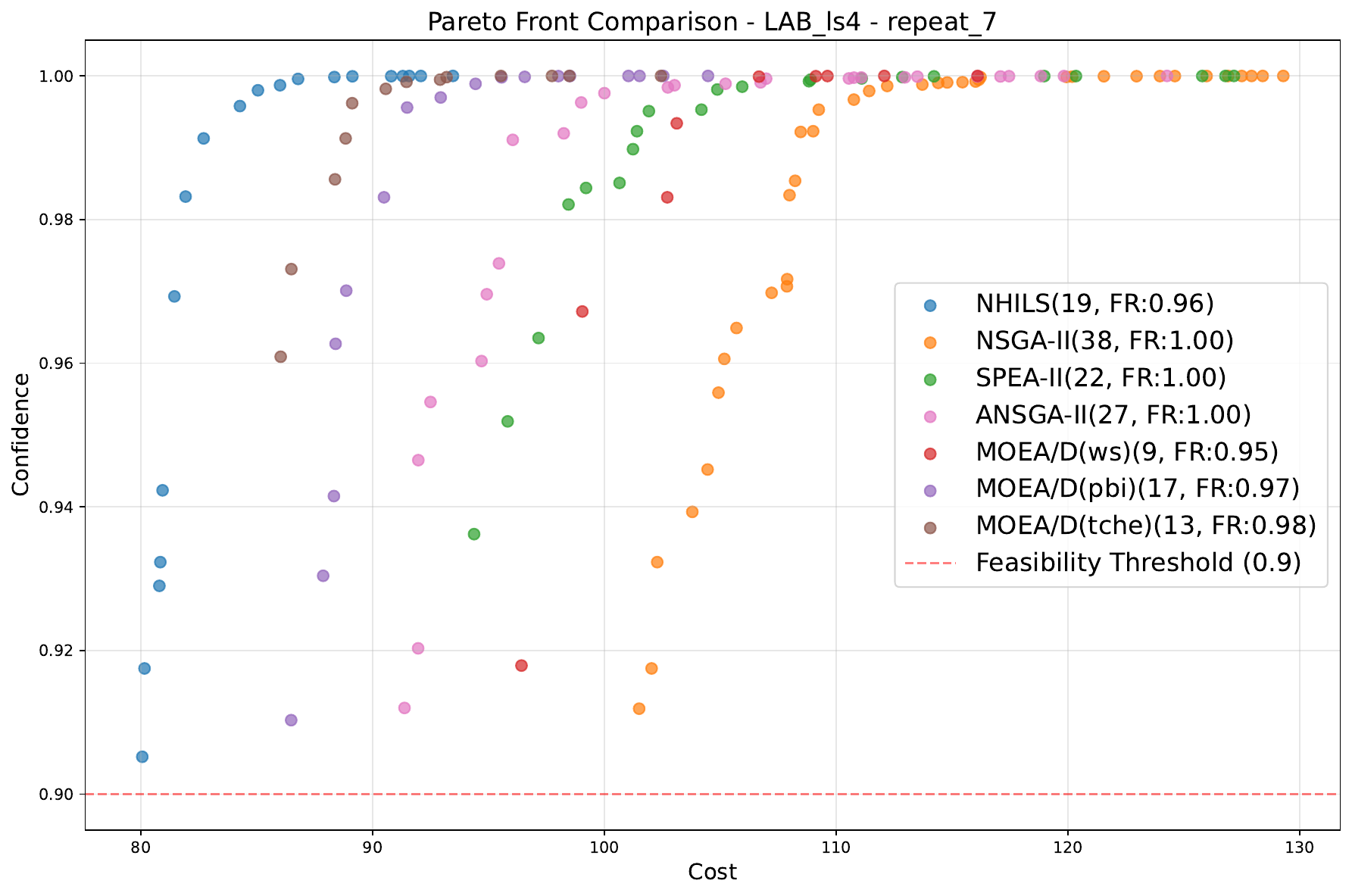}%
			\label{fig:LAB-ls4}}
		\hfil
		\subfloat[LAB-ls5]{\includegraphics[width=0.6\columnwidth]{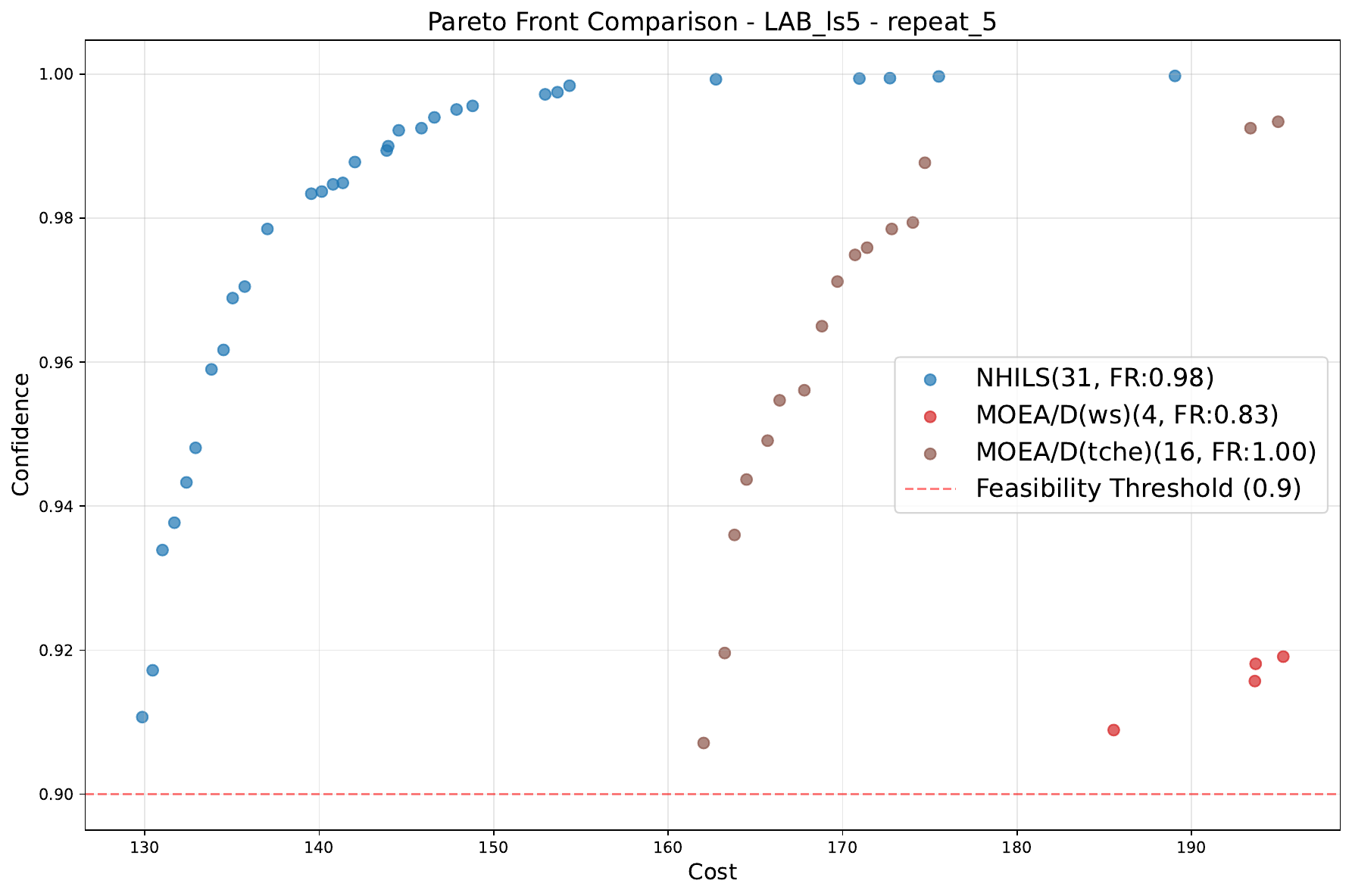}%
			\label{fig:LAB-ls5}}
		\hfil
		\subfloat[LAB-ls6]{\includegraphics[width=0.6\columnwidth]{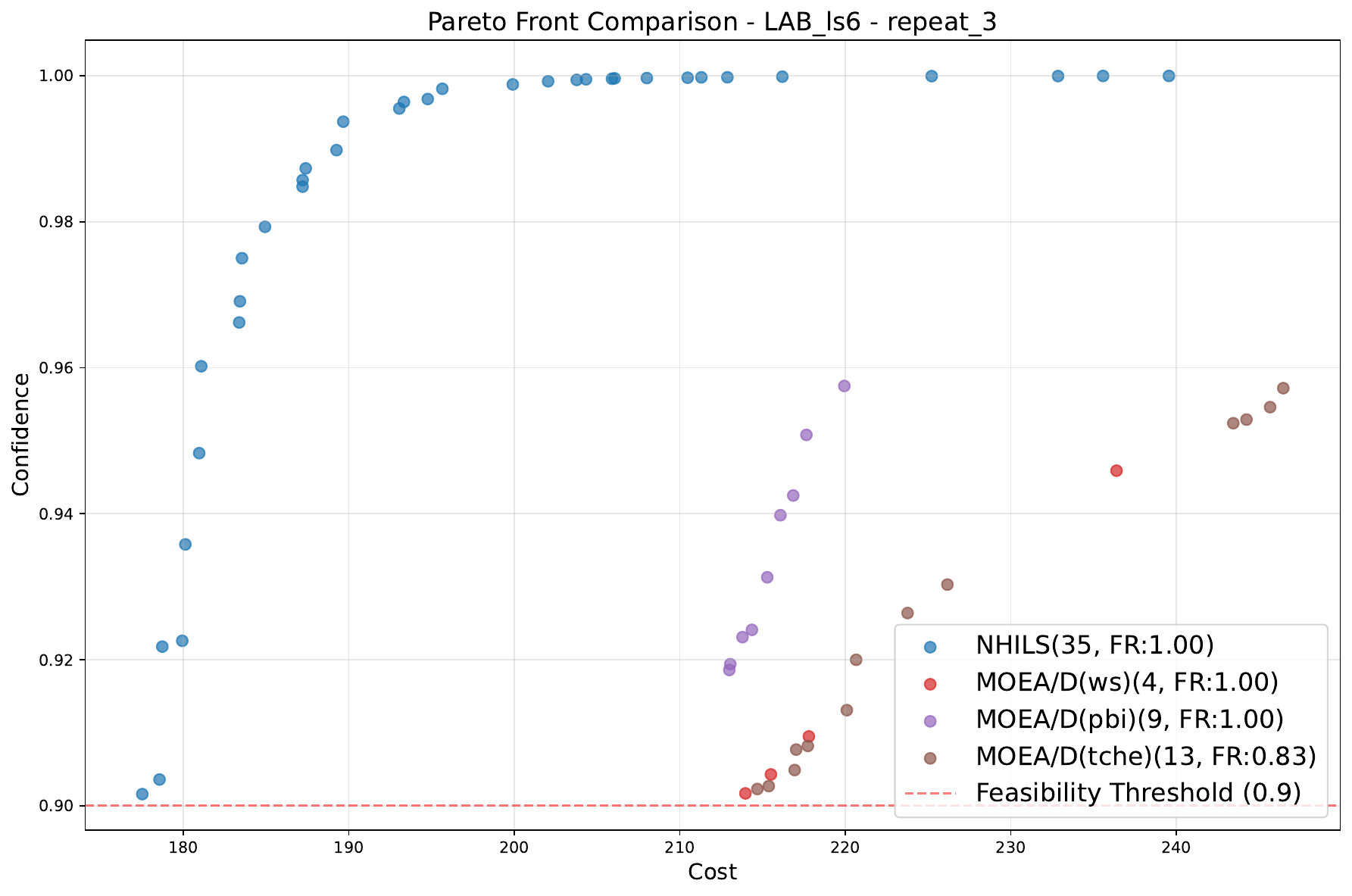}%
			\label{fig:LAB-ls6}}
		
		\caption{Pareto front comparisons of algorithms on LAB instances. From left to right and top to bottom: LAB-ls1 to LAB-ls6.}
		\label{fig:pareto_front_lab}
	
		\centering
		\subfloat[APP-ls1]{\includegraphics[width=0.6\columnwidth]{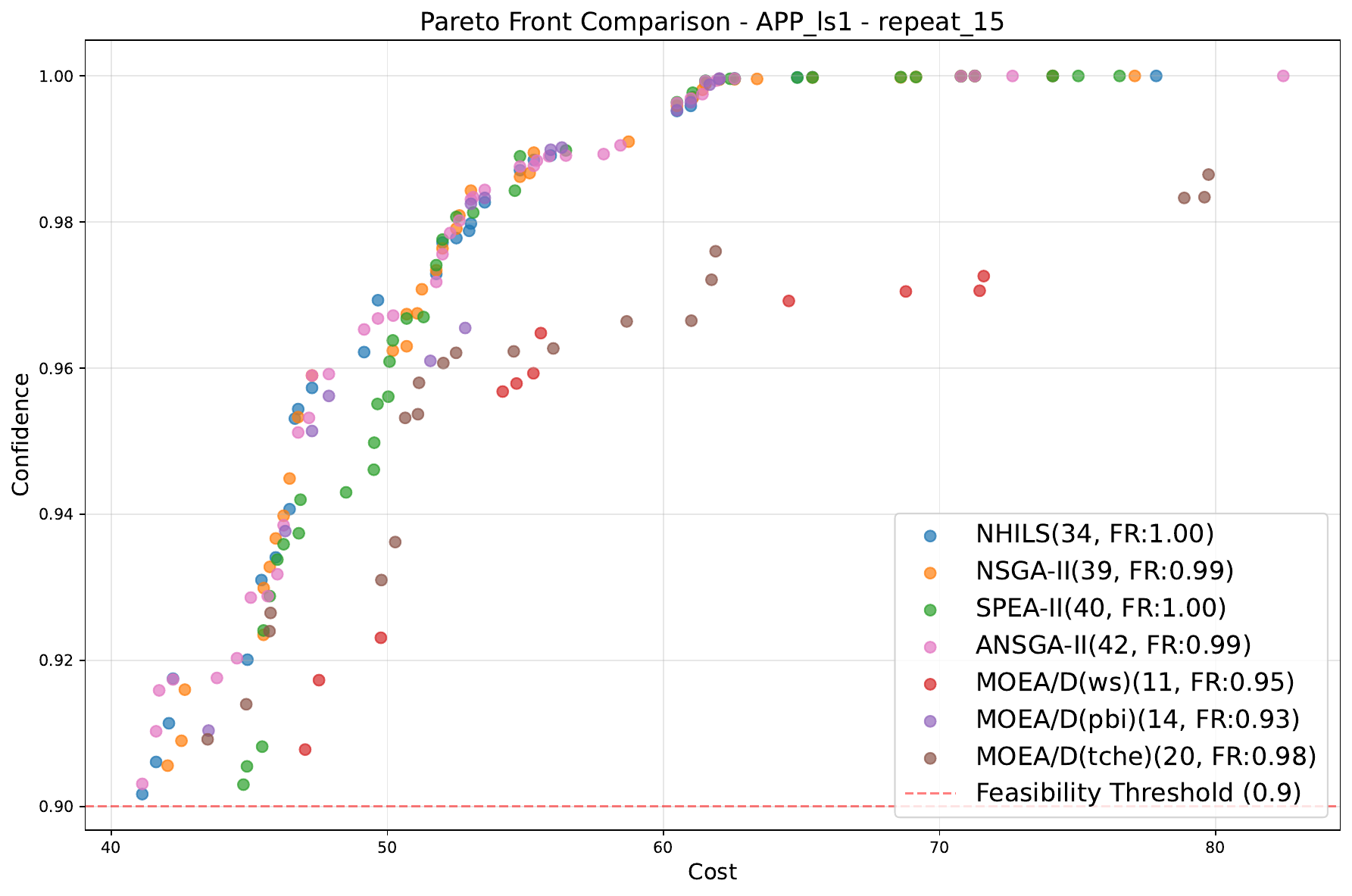}%
			\label{fig:APP-ls1}}
		\hfil
		\subfloat[APP-ls2]{\includegraphics[width=0.6\columnwidth]{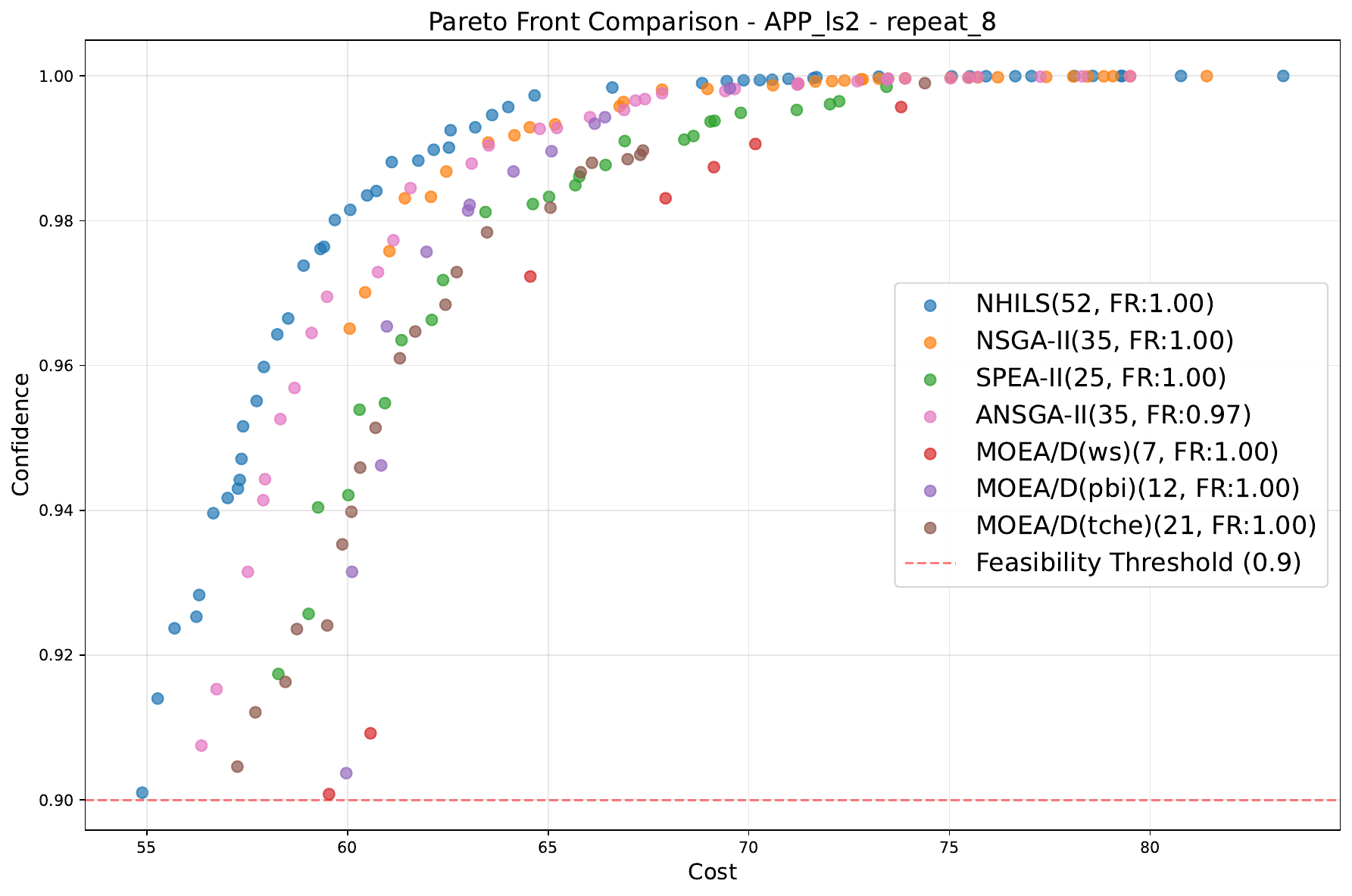}%
			\label{fig:APP-ls2}}
		\hfil
		\subfloat[APP-ls3]{\includegraphics[width=0.6\columnwidth]{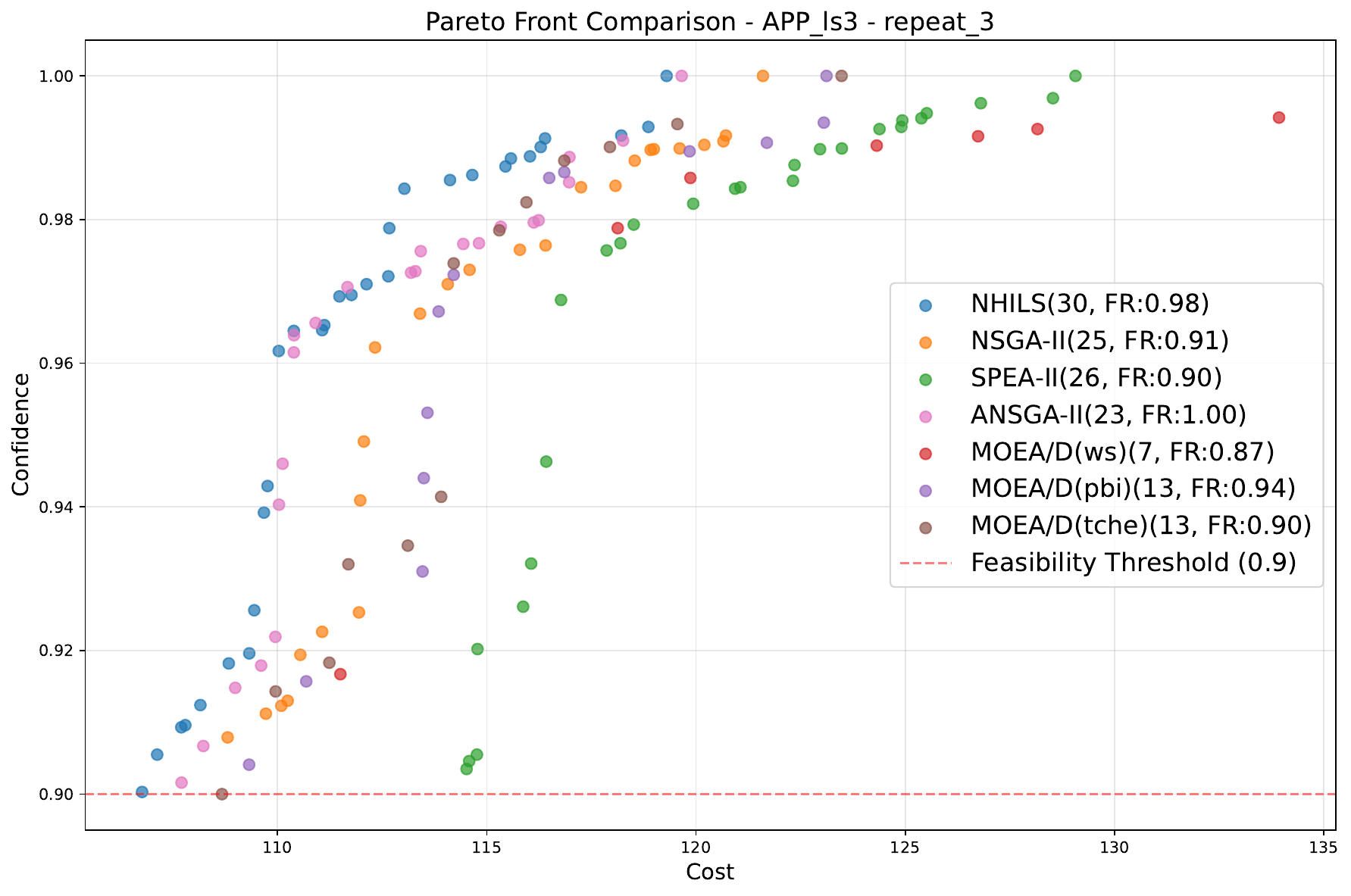}%
			\label{fig:APP-ls3}}
		\hfil
		\subfloat[APP-ls4]{\includegraphics[width=0.6\columnwidth]{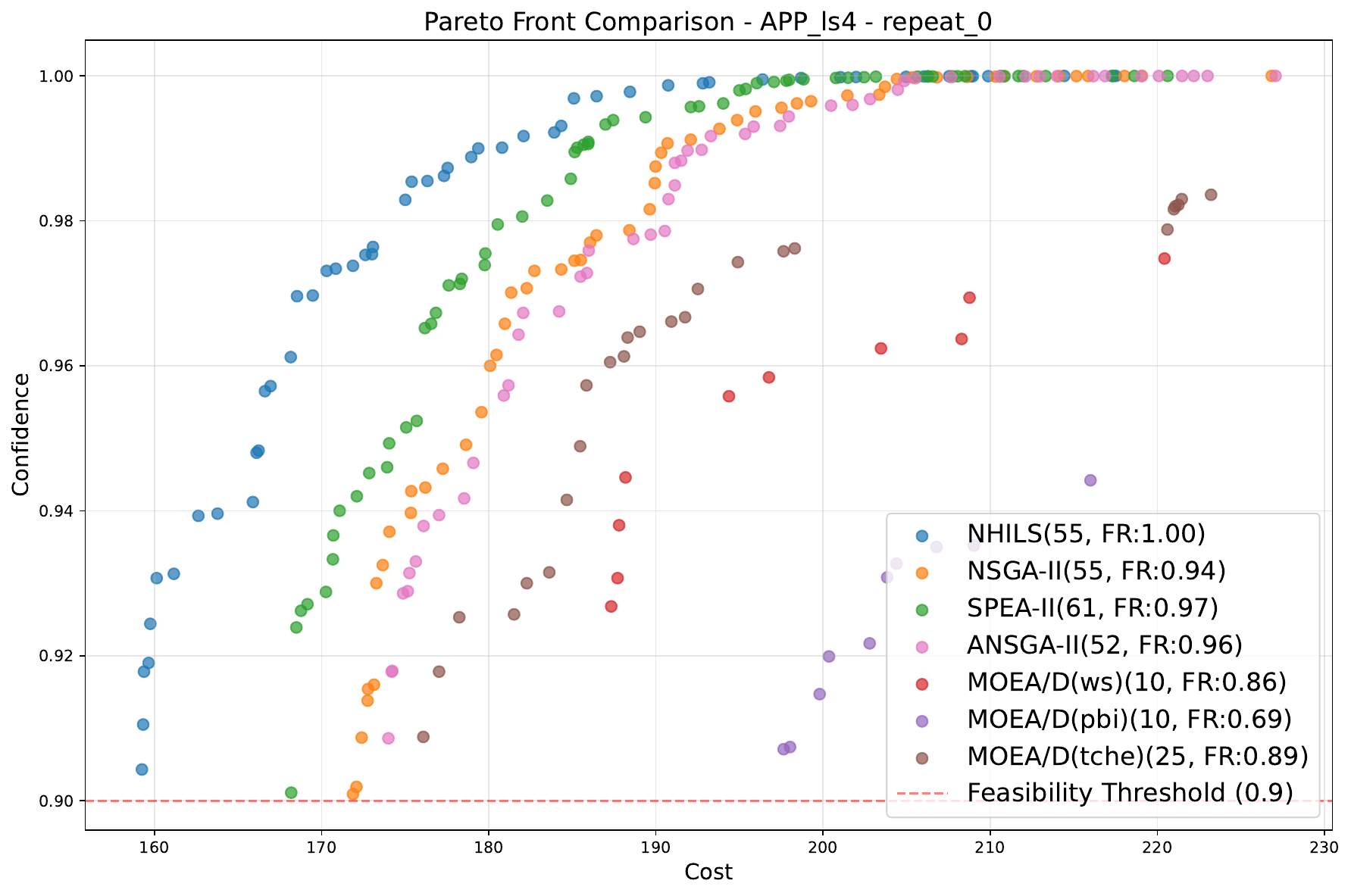}%
			\label{fig:APP-ls4}}
		\hfil
		\subfloat[APP-ls5]{\includegraphics[width=0.6\columnwidth]{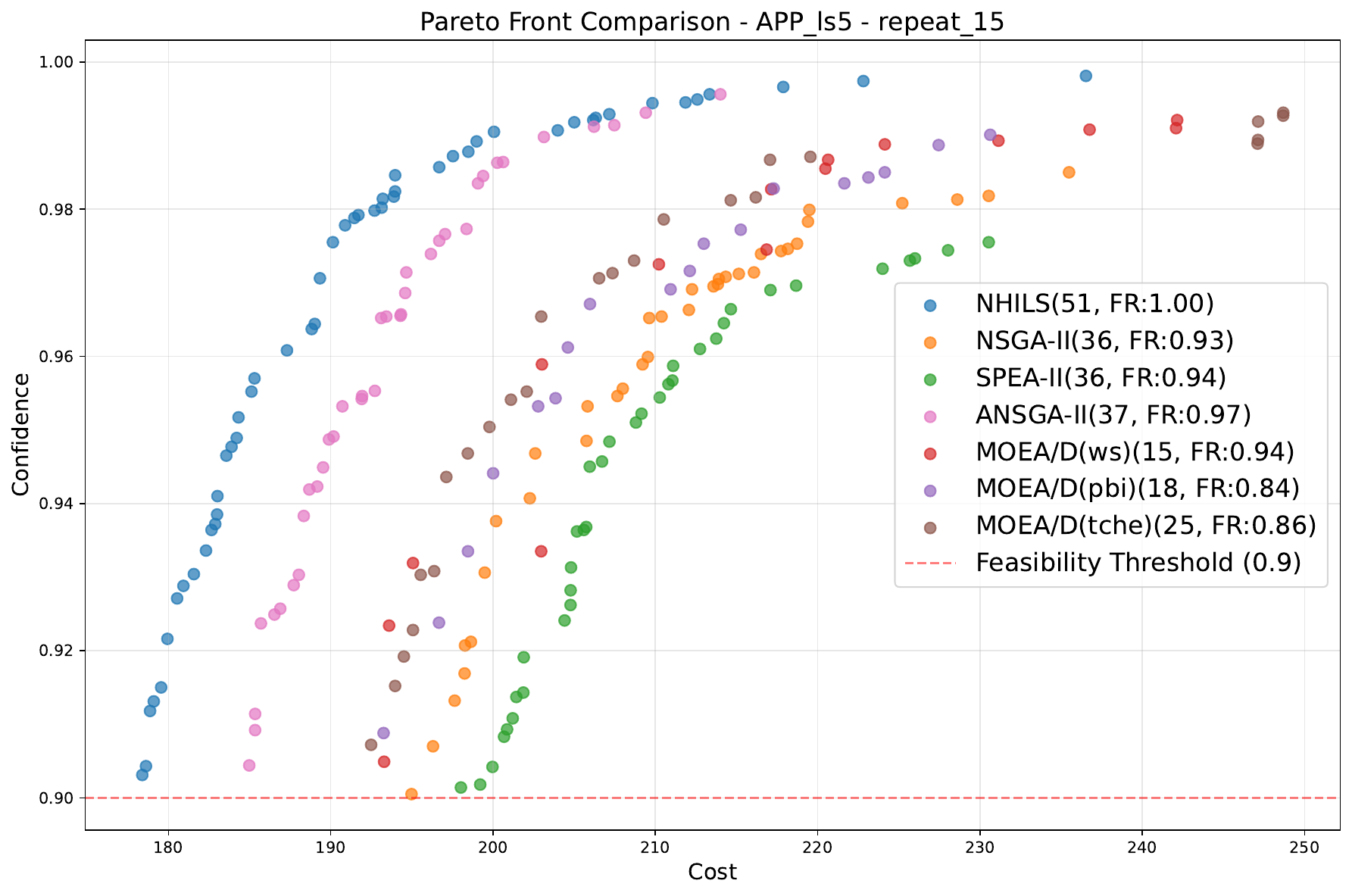}%
			\label{fig:APP-ls5}}
		\hfil
		\subfloat[APP-ls6]{\includegraphics[width=0.6\columnwidth]{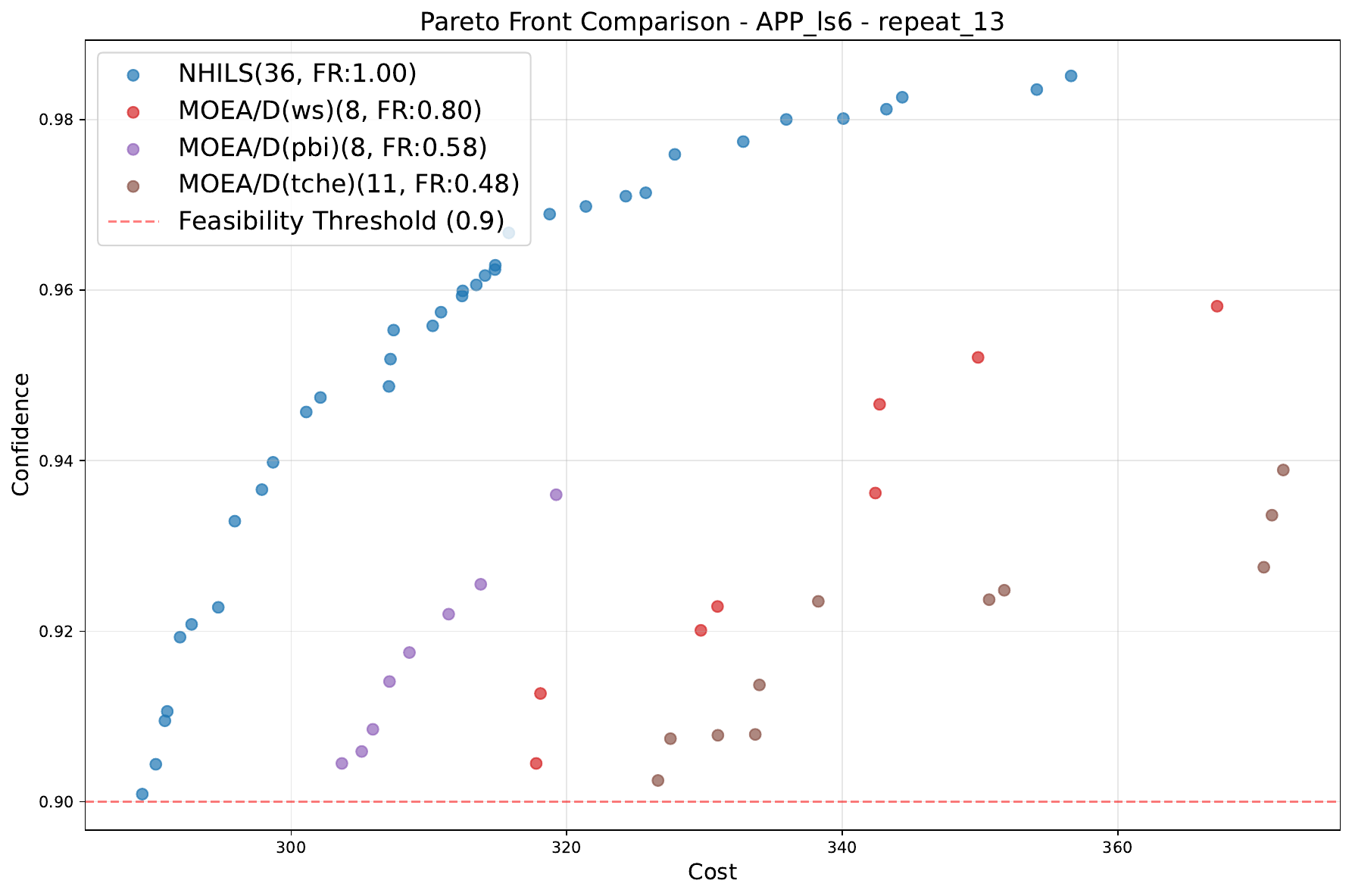}%
			\label{fig:APP-ls6}}
		
		\caption{Pareto front comparisons of algorithms on APP instances. From left to right and top to bottom: APP-ls1 to APP-ls6}
		\label{fig:pareto_front_app}
	\end{figure*}
	
	Figure~\ref{fig:pareto_front_lab} and \ref{fig:pareto_front_app} present the Pareto front comparisons of seven optimization algorithms across twelve test instances with increasing complexity. 
	NHILS demonstrates superior performance in generating well-distributed and converged Pareto fronts.
	It also achieves better boundary coverage and solution uniformity.
	The performance advantage becomes more pronounced with increasing problem scale.
	
	\subsection{Ablation and Efficiency Analysis}
	\subsubsection{Impact of Initialization and Local Search}
	To isolate the contribution of the algorithmic components, all variants in the ablation study (including the standard NSGA-II baseline) utilize the OPERA-MC evaluation approach.
	The results presented in Table \ref{tab:ablation_study} demonstrate the critical contributions of each algorithmic component of NHILS. 
	The complete algorithm (NHILS) achieves superior performance with a hypervolume of 5.831 and IGD of 0.290. 
	The removal of local search results in notable performance degradation, with hypervolume decreasing by 9.1\% to 5.298 and IGD deteriorating dramatically by 565\% to 1.929. 
	The absence of hybrid initialization reduces hypervolume by 27.0\% to 4.258 and worsening IGD by 650\% to 2.177.
	While the complete algorithm necessitates the longest execution time (859.32 seconds), it delivers optimal optimization outcomes that justify the additional computational cost. 
	The standard NSGA-II produces substantially inferior results with hypervolume of only 3.305 and IGD of 2.823 though computationally efficient (20.36 seconds). 
	This performance gap—representing a 76.4\% improvement in hypervolume and 89.7\% improvement in IGD achieved by the full algorithm—validates the effectiveness of our integrated approach. 
	
	\begin{table}[htbp]
		\caption{Ablation Study Results of NHILS Components}
		\label{tab:ablation_study}
		\centering
		\resizebox{\columnwidth}{!}{%
			\begin{tabular}{lccc}
				\toprule
				\textbf{Variant} & \textbf{HV} & \textbf{IGD}  & \textbf{Time (s)} \\
				\midrule
				Complete & $\mathbf{5.83^* \pm 2.88}$ & $\mathbf{0.29^* \pm 0.34}$& 859.32 $\pm$ 873.96 \\
				No Local Search & 5.30 $\pm$ 2.44 & 1.93 $\pm$ 2.07 & 107.32 $\pm$ 100.06 \\
				No HybridInit & 4.26 $\pm$ 1.75 & 2.18 $\pm$ 4.97 & 334.24 $\pm$ 191.82 \\
				NSGA-II & 3.31 $\pm$ 1.20 & 2.82 $\pm$ 3.09 & $\mathbf{20.36^* \pm 15.73}$ \\
				\bottomrule
			\end{tabular}
		}
	\end{table}
	
	\subsubsection{Efficiency of OPERA-MC}
	\label{subsec:monte_carlo}
\begin{figure}[!t]
		\centering
		\includegraphics[width=3in]{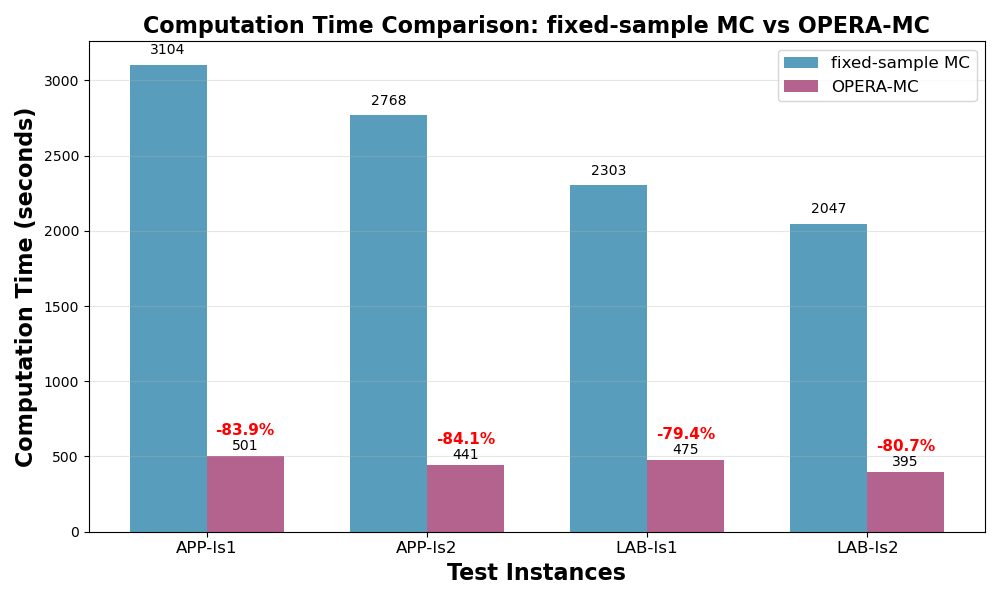}
		\caption{Acceleration performance of OPERA-MC compared with fixed-sample MC on various instances.} 
		\label{chap5:computation_time_comparison}
	\end{figure}

	The proposed OPERA-MC demonstrates exceptional computational efficiency improvements compared to fixed-sample MC (using $10^6$ samples), as illustrated in Figure~\ref{chap5:computation_time_comparison}. 
	Across all test instances, OPERA-MC achieves substantial time reductions ranging from 79.4\% to 83.9\%, with an average reduction of 81.7\%. 
	For the APP-ls1 instance, computation time decreases dramatically from 3104.15 seconds to 500.81 seconds, representing an 83.9\% improvement. 
	For LAB-ls2, the execution time reduces from 2046.53 seconds to 394.93 seconds (80.7\% reduction). 
	These consistent and significant efficiency gains across both APP and LAB instances validate the effectiveness of our multi-stage adaptive sampling strategy in intelligently allocating computational resources. 
	
	\section{Conclusion}
	\label{sec:conclusion}

	This paper introduced an important emerging variant of the knapsack problem: the multi-objective chance-constrained multiple-choice knapsack problem (MO-CCMCKP).
	By addressing cost-reliability trade-offs under implicit uncertainty, this model provides a more realistic framework for decision-making in complex stochastic environments (e.g., 5G network configuration) than traditional deterministic models.
	To solve the unique computational and search challenges of the MO-CCMCKP, we developed the OPERA-MC evaluation method and the NHILS algorithm.
	Experimental results on synthetic benchmarks and real-world application benchmarks demonstrate that our approach significantly reduces simulation overhead, by an average of 81.7\%, while consistently outperforming several state-of-the-art multi-objective baselines in convergence, diversity, and feasibility.
	
	Several future research avenues remain.
	First, the efficiency of OPERA-MC could be further improved by incorporating a learned proxy function, such as a neural network~\cite{LiuZTY23}, to predict solution quality based on historical simulation data and further refine sampling patterns.
	Second, while our current model assumes mutually independent item weights, future work should address the complex inter-item dependencies often found in domains like finance~\cite{11277388}.
	Finally, rather than relying on hand-designed initialization and local search operators, future improvements could leverage general-purpose pretrained models to automatically generate them~\cite{LiuCQ0O24}, potentially identifying high-quality solutions more effectively in high-dimensional combinatorial landscapes.

	\appendices
	
	\section{A Theoretical Quantitative Analysis between Estimation Error and Sample Size}
	To theoretically quantify the relationship between estimation error and sample size of data, we construct a new random variable $S$ that is the sum of multiple random variables such that the variable itself is a random variable. 
	Then we define a Bernoulli variable $Z\in \{0,1\}$, i.e.,
	\begin{equation}
		Z=\left\{
		\begin{aligned}
			1, &\quad\quad S\leq W \\
			0, &\quad\quad S> W
		\end{aligned}
		,\right.
	\end{equation}
	for a combination of items with $L$ sample size, where $p$ is the true confidence level, i.e., the expectation $\mathbb{E}Z$ of the variable $Z$. 
	Hoeffding's inequality shows that for independent identically distributed random variables $Z_1,Z_2,\ldots,Z_L$ taking values $[a,b]$, for any $\epsilon>0$, the following inequality holds,
	\begin{equation}
		Pr(|\mathbb{E}[Z]-\frac{1}{L}\sum_{l=1}^L Z_l|\geq \epsilon)\leq 2exp(-\frac{2L\epsilon^2}{(b-a)^2}).
	\end{equation}
	For our problem, we are concerned with the true value is lower than the estimate and greater than the error, i.e., where $\epsilon$ is the estimation error corresponding to the confidence level. 
	For the Bernoulli distribution, we have $\mathbb{E}Z=p$ and $a=0,b=1$. 
	Thus, the inequality becomes
	\begin{equation}
		Pr(p-f(Z_1,Z_2,\ldots,Z_L)\leq -\epsilon)\leq exp(-2L\epsilon^2),
	\end{equation}
	where $f(Z_1,Z_2,\ldots,Z_L)$ denotes the confidence evaluation algorithm on some combination of items. 
	Assuming that at least 5 of the 10 solutions in the output solution set are true feasible solutions, and $f(Z_1,Z_2,\ldots,Z_L)$ is accurate, then we have $1-exp(-2L\epsilon^2)\geq 5/10$. 
	Finally, it leads to 
	\begin{equation}
		L\geq \frac{ln2}{2\epsilon^2}
	\end{equation}
	
\begin{figure}[htpb]
		\centering
		\includegraphics[width=0.8\columnwidth]{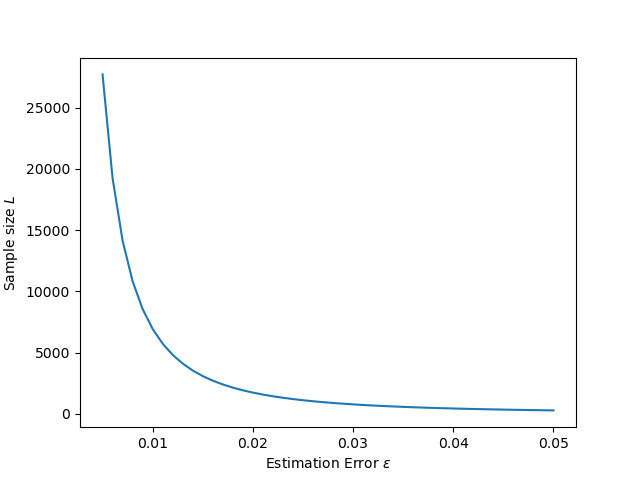}
		\caption{Sample size requirement with different error in theory} \label{fig:samplesize_curve}
	\end{figure}

	As this bound is derived from Hoeffding's inequality, Chernoff bound is proposed in \cite{tempo1996probabilistic} and a more tight one is proposed in \cite{alamo2010sample}. 
	However, all of these inequalities indicate that a large amount of samples is necessary. 
	The curve of the variation of $L$ with $\epsilon$ is shown in Figure \ref{fig:samplesize_curve}.
	The minimum sample sizes required for different estimation error are listed in Table \ref{tab:samplesize}. 
	
	Overall, Hoeffding provides the tightest bounds, especially suitable for larger errors. 
	However, as error requirements become smaller, the bounds provided by Alamo are more advantageous. 
	Practically, it is verified that the amount of data required to meet the corresponding error requirement for the true confidence will be much smaller than the theoretical value obtained.
	
	\begin{table}[!t]
		\centering
		\caption{Relationship between EE and sample size}
		\begin{tabular}{ccccc}
			\toprule
			\bfseries Estimation error($\%$) & 5 & 0.5 & 0.05 & 0.005 \\
			\midrule
			{\bfseries Hoeffding's bound} &139 &13863 &1386295 & 138629437\\
			{\bfseries Chernoff's bound \cite{tempo1996probabilistic}} &278 &27726 &2772589 &277258873\\
			{\bfseries Alamo's bound \cite{alamo2010sample}} &1573 &15723 &157227 &1572269\\
			\bottomrule
			\label{tab:samplesize}
		\end{tabular}
	\end{table}
	
	\section{Complete Derivation of the Order‑Preservation Bound}
	\label{app:proof}
	
	We provide a self‑contained derivation of Theorem 4.1 using Hoeffding’s inequality.
	Let \(X_1,\dots,X_{N_A} \sim \mathrm{Bernoulli}(p_A)\) and \(Y_1,\dots,Y_{N_B} \sim \mathrm{Bernoulli}(p_B)\) be independent samples. 
	Define the empirical means $\hat{P}_A = \frac{1}{N_A}\sum_{i=1}^{N_A} X_i$ and $\hat{P}_B = \frac{1}{N_B}\sum_{j=1}^{N_B} Y_j$.
	Consider the random variable \(Z = \hat{P}_B - \hat{P}_A\). 
	Its expectation is
	\begin{equation}
		\mathbb{E}[Z] = p_B - p_A = -\Delta,\Delta = p_A - p_B > 0.
	\end{equation}
	
	We rewrite \(Z\) as a sum of \(N_A+N_B\) independent terms:
	\begin{equation}
		Z = \sum_{j=1}^{N_B} \frac{Y_j}{N_B} \;+\; \sum_{i=1}^{N_A} \left(-\frac{X_i}{N_A}\right).
	\end{equation}
	Each term \(\frac{Y_j}{N_B}\) lies in \([0, \frac{1}{N_B}]\). And each term \(-\frac{X_i}{N_A}\) lies in \([-\frac{1}{N_A}, 0]\). 
	The total squared range for Hoeffding’s inequality is therefore
	\begin{equation}
		\sum_{k=1}^{N_A+N_B} (\text{range}_k)^2 
		= N_B\left(\frac{1}{N_B}\right)^2 + N_A\left(\frac{1}{N_A}\right)^2 
		= \frac{1}{N_B} + \frac{1}{N_A}.
	\end{equation}
	
	Applying Hoeffding’s inequality with deviation \(t = \Delta\) gives
	\begin{equation}
		\mathbb{P}(Z \ge 0) 
		= \mathbb{P}\!\bigl(Z - \mathbb{E}[Z] \ge \Delta\bigr)
		\le \exp\!\left( -\frac{2\Delta^2}{\frac{1}{N_B} + \frac{1}{N_A}} \right).
	\end{equation}
	
	Since the multi‑stage algorithm ensures \(N_A \ge N_B\) for any pair where \(S_A\) is promoted further than \(S_B\), we have \(\frac{1}{N_A} \le \frac{1}{N_B}\). 
	Hence,
	\begin{equation}
		\frac{1}{N_B} + \frac{1}{N_A} \le \frac{2}{N_B},
	\end{equation}
	which, when substituted into the exponent, yields the simpler bound
	\begin{equation}
		\mathbb{P}(\hat{P}_B \ge \hat{P}_A) \le \exp\!\bigl( - N_B \Delta^2 \bigr).
	\end{equation}
	
	\ifCLASSOPTIONcaptionsoff
	\newpage
	\fi

	\bibliographystyle{IEEEtran}
	\bibliography{ref.bib}
	
\end{document}